\documentclass[square]{article}

\RequirePackage[OT1]{fontenc}
\RequirePackage[authoryear, square]{natbib}
\RequirePackage[colorlinks,citecolor=blue,urlcolor=blue]{hyperref}
\RequirePackage{hypernat}
\usepackage[sc]{mathpazo} 
\usepackage[T1]{fontenc} 
\usepackage{microtype} 

\usepackage{titling} 

\usepackage[bottom]{footmisc}
\usepackage{color,colortbl}
\usepackage{float}
\usepackage{graphicx}
\usepackage{amsmath}
\usepackage{amssymb}
\usepackage{amsfonts}
\usepackage{amsthm}
\usepackage{thmtools}
\usepackage{subcaption}
\usepackage{tikz}
\usetikzlibrary {positioning}
\usetikzlibrary{fit}
\usepackage{csquotes}
\usepackage{cleveref}
\usepackage[noend]{algpseudocode}
\usepackage{algorithm}
\usepackage{mathtools}
\usepackage[leftcaption]{sidecap}
\usepackage{stackengine}
\usepackage{xcolor}
\definecolor{light-gray}{gray}{0.75}
\usepackage{fancyhdr} 
\usepackage[hmarginratio=1:1,top=32mm]{geometry} 

\definecolor{LightCyan}{rgb}{0.75,1,0.75}
\definecolor{Light2}{rgb}{1,0.75,0.75}
\definecolor{Light3}{rgb}{0.75,0.75,1}
\definecolor{Light4}{rgb}{0.75,0.75,0.75}
\usetikzlibrary{arrows}
\usetikzlibrary{backgrounds}
\definecolor{graphNode1}{rgb}{0.267,0.004,0.329}
\definecolor{graphNode2}{rgb}{0.267,0.227,0.514}
\definecolor{graphNode3}{rgb}{0.192,0.408,0.557}
\definecolor{graphNode4}{rgb}{0.129,0.565,0.549}
\definecolor{graphNode5}{rgb}{0.208,0.718,0.475}
\definecolor{graphNode6}{rgb}{0.561,0.843,0.267}
\definecolor{graphNode7}{rgb}{0.992,0.906,0.145}
\tikzset{
  treenode/.style = {shape=rectangle, rounded corners,  draw, align=center},
  arn_n/.style = {treenode, top color=white, bottom color=gray!20},
  arn_n1/.style = {treenode, top color=white, bottom color=graphNode1},
  arn_n2/.style = {treenode, top color=white, bottom color=graphNode2},
  arn_n3/.style = {treenode, top color=white, bottom color=graphNode3},
  arn_n4/.style = {treenode, top color=white, bottom color=graphNode4},
  arn_n5/.style = {treenode, top color=white, bottom color=graphNode5},
  arn_n6/.style = {treenode, top color=white, bottom color=graphNode6},
  arn_n7/.style = {treenode, top color=white, bottom color=graphNode7}
}

\definecolor{LightCyan}{rgb}{0.75,1,0.75}
\definecolor{Light2}{rgb}{1,0.75,0.75}
 \definecolor{Light3}{rgb}{0.75,0.75,1}

\newcommand{\pkg}[1]{{\fontseries{b}\selectfont #1}}

\DeclarePairedDelimiter\abs{\lvert}{\rvert}%

\def\CC{{C\nolinebreak[4]\hspace{-.05em}\raisebox{.4ex}{\tiny\bf ++}}}

\newtheorem{lem}{Lemma}
\newtheorem{thm}{Theorem}

\newtheorem{defn}{Definition}
\newlength\mytemp
\newlength\myoffset
\newcommand\colblock[4][blue!20]{%
    \setlength\mytemp{#2\baselineskip}%
    \setlength\mytemp{.5\mytemp-\myoffset}%
    \belowbaseline[0pt]{%
      \fboxsep=-\fboxrule\fbox{\colorbox{#1}{\rule[-\mytemp]{0ex}{#2\baselineskip}%
          \makebox[#3\baselineskip]{$#4$}}}}}

\title{Selective Clustering Annotated using Modes of Projections} 
\author{%
\textsc{Evan Greene}\thanks{Corresponding author} \\[1ex] 
\normalsize \href{mailto:egreene@fredhutch.org}{egreene@fredhutch.org} 
\and 
\textsc{Greg Finak} \\[1ex] 
\normalsize \href{mailto:gfinak@fredhutch.org}{gfinak@fredhutch.org} 
\and 
\textsc{Raphael Gottardo}\thanks{Corresponding author} \\[1ex] 
\normalsize \href{mailto:rgottard@fredhutch.org}{rgottard@fredhutch.org}\\
\\
\normalsize Vaccine and Infectious Disease Division\\
\normalsize Fred Hutchinson Cancer Research Center\\
\normalsize  Seattle, WA 98109, USA \\
}
\date{} 
\renewcommand{\maketitlehookd}{%
\begin{abstract}
\noindent 
 Selective clustering annotated using modes of projections (SCAMP) is a new clustering algorithm for data in $\mathbb{R}^p$. SCAMP is motivated from the point of view of non-parametric mixture modeling. Rather than maximizing a classification likelihood to determine cluster assignments, SCAMP casts clustering as a search and selection problem. One consequence of this problem formulation is that the number of clusters is \textbf{not} a SCAMP tuning parameter.
  \linebreak
  \linebreak
  The search phase of SCAMP consists of finding sub-collections of the data matrix, called candidate clusters, that obey shape constraints along each coordinate projection. 
  An extension of the dip test [\cite{hartigan1985dip}] is developed to assist the search.
  Selection occurs by scoring each candidate cluster with a preference function that quantifies prior belief about the mixture composition.
  Clustering proceeds by selecting candidates to maximize their total preference score. 
  SCAMP concludes by annotating each selected cluster with labels that describe how cluster-level statistics compare to certain dataset-level quantities.
  \linebreak
  \linebreak
  SCAMP can be run multiple times on a single data matrix. 
  Comparison of annotations obtained across iterations provides a measure of clustering uncertainty.
  Simulation studies and applications to real data are considered.
  A \CC \ implementation with R interface is available at \href{https://github.com/RGLab/scamp}{https://github.com/RGLab/scamp}.
\end{abstract}
}
\begin{document}

\maketitle

\bigskip

\section{Introduction}\label{section:introduction}

This paper introduces a new algorithm to cluster data in $\mathbb{R}^p$ called selective clustering annotated using modes of projections (SCAMP).
Clustering is a common task in data analysis.
Numerous approaches to clustering have been developed over the past century: see 
\cite{cormack1971review}, \cite{jain2010data}, and \cite{hennig2015handbook} for 
descriptions of many such methods.
We begin by giving an overview of the SCAMP algorithm, in which we relate it to prior 
work in the clustering literature.

The SCAMP algorithm is inspired by both the mixture modeling 
and density based approach to clustering.
Suppose the rows $x_i = (x_{i,1},\ldots,x_{i,p})$ of a data matrix $X_{k\times p}$  
are samples from a distribution $F$ with distribution function (df)
\[
  F(x_i) = \sum_{m=1}^g \pi_m F_m(x_i) \text{ with } 0 \leq \pi_m \leq 1\ \ \forall\ \ m \text{, and } \sum_{m=1}^g \pi_m = 1\ .
\]
In mixture modeling, each component df $F_m$ of the mixture is typically associated 
with an observed cluster.
To cluster the data, a parametric model, such as the multivariate normal 
[\cite{fraley2002model}], uniform normal [\cite{banfield1993model}], multivariate t 
[\cite{peel2000robust}],
or a skew variant of one of these distributions [\cite{lee2013model}], 
is typically assumed to describe the component distributions $F_m$.
Non-parametric approaches are also possible: \cite{li2007nonparametric} suppose the 
mixture is made up of non-parametric components and estimates them with Gaussian 
kernels;
\cite{rodriguez2014univariate} and \cite{paez2017modeling} take a Bayesian approach to 
estimating mixture models based on unimodal distributions; 
\cite{kosmidis2016model} take a copula-based approach that influenced the model formulation 
used in this work.
The EM algorithm of \cite{dempster1977maximum} (or one of its extensions) is then used 
to find local maximizers of the likelihood for a fixed number of clusters $g$.
The number of components $g^*$ can be determined in terms of the model by picking the $g^*$ that optimizes some criterion (such as BIC) over a user-specified range of $g$, and subsequently refined by merging clusters with, for example, the method of \cite{baudry2010combining} or \cite{tantrum2003assessment}.
The review by \cite{mcnicholas2016model} provides a detailed overview of this approach and references to many recent developments.

SCAMP takes a perspective similar to \cite{paez2017modeling}: clusters are defined in terms of unimodality.
In many scientific contexts, unimodality of observations along a measurement coordinate of the data matrix $X$ reflects physical homogeneity of interest.
Using Sklar's theorem, (see Theorem 2.10.9, \cite[pg. 46]{nelsen2007introduction}, as well as  \cite{kosmidis2016model}), 
we can formalize a data model: we assume an observation $x$ is sampled from the mixture with df
\begin{align}
  F(x_i) = \sum_{m=1}^g \pi_m C_m\left(F_{m,1}(x_{i,1}),F_{m,2}(x_{i,2}),\ldots,F_{m,p}(x_{i,p})\right)\ , \label{scamp:cluster_distribution}
\end{align}
where each $C_m$ is a $p$-copula for the $m^{th}$ component  subject to the constraint that $F_{m,j}(x_{i,j})$ 
is unimodal for all $1 \leq m \leq g$ and $1 \leq j \leq p$.

We will reference this formalization \eqref{scamp:cluster_distribution} as we discuss SCAMP's design choices.
However, SCAMP does not attempt to estimate model parameters using an EM-based approach.
Indeed, without further assumptions on the copulas $C_m$, using the mixture $F$ to cluster the data is a non-parametric problem.
A standard approach to non-parametric clustering is to estimate the level-sets of $F$ at some level $\lambda > 0$ of the mixture df $F(x)$ (see figure \ref{fig:excessmassfail}).
Clustering is then achieved by associating observations with local modes.

\cite{hartigan1975clustering} provides an early discussion of how level sets can be used in clustering.
In it, Hartigan noted that different choices of $\lambda$ induce a hierarchical structure, now commonly called a cluster tree.
He later investigated the problem of  estimating level sets in the two-dimensional case, with an aim of testing for bimodality [\cite{hartigan1987estimation}].
\cite{muller1987using,muller1991excess} also studied this testing problem, leading to the development of the excess mass test.
\cite{polonik1995measuring,polonik1998silhouette} studied theoretical properties of the excess mass, and later the silhouette.
\cite{stuetzle2003estimating} developed a method called runt pruning, which aims to construct a pruned cluster tree using a nearest neighbor density estimate.
\cite{stuetzle2010generalized} later generalized the pruning method in a graph based clustering approach.
Analysis of the stability of both level set of cluster tree estimates are discussed in [\cite{rinaldo2012stability}].
Two consistent procedures, the first which generalizes single-linkage, the second based on the k-nearest neighbor graph, are analyzed in [\cite{chaudhuri2014consistent}].
More recently in \cite{chen2017density}, methods for constructing confidence sets for level sets by \cite{jankowski2012confidence} and \cite{mammen2013confidence} 
are compared to a new bootstrap based approach.

The level-set approach to clustering makes no distributional assumptions and generalizes naturally to arbitrary dimension.
Compared to mixture modeling, this provides robustness in the event the component distributions of the mixture are specified incorrectly.
On the other hand, approaches that rely on non-parametric density estimators suffer the curse of dimensionality [\cite{nagler2016evading}].
For example, \cite{stuetzle2010generalized} observe that it is infeasible to use a plug-in binned density-estimator to estimate a cluster tree for a data matrix with ten features since ``ten bins per variable in ten dimension would result in $10^{10}$ bins''.
Bandwidth selection also affects the modal structure of the estimated density, which adds an additional complication to non-parametric modal clustering.

The SCAMP algorithm attempts to cluster a data matrix by associating observations with 
components of the mixture distribution \eqref{scamp:cluster_distribution}.   
In deriving these associations, it tries to reap some of the robustness benefits of 
non-parametric modal clustering without suffering excessively from the curse of 
dimensionality. Approaches that combine aspects of non-parametric modal 
clustering and mixture modeling have been proposed by \cite{tantrum2003assessment},
\cite{hennig2010methods}, \cite{scrucca2016identifying}, and \cite{chacon2016mixture}.
Broadly speaking, these methods rely on fitting a mixture model (typically Gaussian)
for a fixed number of  components, and then using density considerations 
to merge components.

Here, SCAMP takes a partitional approach 
that rests on a severe simplification: the SCAMP algorithm only uses     
one-dimensional density estimates when clustering a dataset.
This simplification carries large costs of its own. 
Most notably, higher-dimensional modal structures that are not axis-aligned cannot 
be detected by SCAMP directly.
However, we argue that the benefits of this simplification outweigh the costs, since it
gives rise to many practical consequences.
For example, SCAMP can cluster datasets with a large number of observations without
sub-sampling, since $O(n)$ implementations exist for many of its constituent
algorithms. We will also argue that SCAMP's tuning parameters are relatively easy 
to understand, as they describe characteristics of one-dimensional densities.
In addition, we note SCAMP can be combined with other methods, such as PCA, to find
structures that are not initially evident along the coordinates of
the data matrix. This is discussed in section \ref{section:casestudes} in more depth. 

Underlying the SCAMP algorithm is the assumption that observed data have been generated
by a specific realization of the mixture distribution \eqref{scamp:cluster_distribution}.
Its clustering strategy is based on the following observation:  in a large data matrix $X$, samples from a component of \eqref{scamp:cluster_distribution}
with a large mixing weight $\pi_m$ will produce a sub-collection of rows of $X$ with unimodal empirical distributions along each coordinate projection.
So, to cluster the data matrix, SCAMP begins by searching for such sub-collections of rows of $X$ with unimodal empirical distributions along each coordinate.
We will call the unimodal sub-collections sought by SCAMP $\alpha$-$m$-clusters. We pause to define them:

\begin{defn}
Denote a random subset of the indices $1,2,\ldots,n$ by 
\[
\mathcal{I}_{m}^n \equiv \left\{ \left(i_{(1)},\ldots,i_{(j)}\right) \ \vline \  m \leq j \leq n \text { and } 1 \leq i_{(1)} < i_{(2)} < \ldots < i_{(j)} \leq  n\right\}\ .
\]
When both $m$ and $n$ are understood, we write $\mathcal{I}$ in place of $\mathcal{I}_m^n$.
\end{defn}

\begin{defn}\label{section1:def:amcluster}
  An \textbf{$\alpha$-$m$-cluster} of the matrix $X_{n\times p}$ is any sub-matrix $X_{\mathcal{I}_m^n\times p} \equiv X_{\mathcal{I}\times p}$ such that each of the $p$ coordinate projections of the sub-matrix
  have p-values greater than $\alpha$ when the dip test statistic of \cite{hartigan1985dip} is computed.
\end{defn}

This definition assumes familiarity with the dip test.
A detailed discussion of the dip test occurs later in section \ref{section:doubledip}, 
and so we defer elaboration until then.
We note here only that the methods of \cite{tantrum2003assessment} and 
\cite{hennig2010methods} both rely on the dip test in their cluster merging methods.

The search for $\alpha$-$m$-clusters starts in a similar fashion to that taken by \cite{chan2010using}:
each coordinate of $X$ is tested for multimodality using the dip test.
If the dip test p-value for a coordinate is smaller than the user selected $\alpha$, SCAMP then splits that coordinate into modal sub-collections.
In most cases, SCAMP uses the taut string [\cite{davies2001local,davies2004densities}] to determine the split points (section \ref{section:scamp} provides details on how split points are determined in small samples).
By using the dip to guide the search for $\alpha$-$m$-clusters, SCAMP's decision to split a coordinate into sub-collections has no dependence on bandwidth: 
so as long as the null hypothesis of unimodality is rejected at the user specified level $\alpha$, the search for $\alpha$-$m$-clusters will continue.

The procedure continues recursively on each modal sub-collection of each multimodal coordinate, until SCAMP encounters a sub-collection of observations that meet one of two stopping conditions:
either the sub-collection satisfies the definition of an $\alpha$-$m$-cluster, or the sub-collection contains fewer than $m$ observations.
All branches of a tree that terminate at a leaf with fewer than $m$ observations are pruned.
Notice that after the first recursive step, the density estimates produced by SCAMP are conditional densities that depend on the sequence of modal groupings leading from the root of the data matrix to the specific sub-collection of observations under consideration.
Notice too that a single coordinate of the data matrix can appear multiple times in this search, so long as the multimodality is detected at a particular depth.
In practice we usually constrain the search to split a coordinate only once along any path from a leaf to the root of a tree, both to facilitate cluster interpretation and reduce computational cost.
 
This search generates a forest we call the \textit{annotation forest}. Each tree in the annotation forest satisfies the following conditions:
\begin{enumerate}
  \item{Each tree starts by splitting a coordinate of the data matrix that exhibits multimodality at level $\alpha$ in the root population.}
  \item{Each leaf of each tree in the annotation forest is an $\alpha$-$m$-cluster.}
  \item{Each tree in the forest can be used to define partitional clustering of the data matrix by viewing the leaves of a given tree as the clusters, 
      and combining any observations corresponding to pruned leaves together into a single residual cluster.}\label{section1:cond3}
  \item{Each observation in the data matrix appears in at most one leaf of each tree of the annotation forest.}\label{section1:cond4}
\end{enumerate}
Condition \ref{section1:cond3} can be interpreted as stating that the SCAMP search defines multiple hierarchical clusterings of the data matrix.
Because of this, we call the trees that make up the annotation forest \textit{partition trees}.
Unlike agglomerative methods such as single-linkage clustering 
[\cite{sibson1973slink,hartigan1981consistency}], SCAMP neither requires the specification of nor 
the computation of pair-wise similarities.
While SCAMP has some traits in common with density-based hierarchical clustering approaches such as
DBSCAN [\cite{ester1996density,sander1998density}], OptiGrid [\cite{hinneburg1999optimal}], and 
OPTICS [\cite{ankerst1999optics}],
we note that the SCAMP search does not require the selection of a metric (nor a distance threshold
$\epsilon$),
does not attempt to order its search according to a best split,
and its stopping condition has no dependence on a bandwidth parameter. 
We refer the reader to the discussion on DBSCAN and OPTICS given by \cite{stuetzle2010generalized}.

We will call the $\alpha$-$m$-clusters appearing as leaves in this forest as \textit{candidate clusters}.
Recalling the underlying data model \eqref{scamp:cluster_distribution},  we view each candidate cluster as a sample from some component $C_m$ of the mixture df $F$.
However, condition \ref{section1:cond4} implies SCAMP must select a subset of the candidate clusters to partition the dataset.
This is because, by construction, a single row of the data matrix appears in multiple candidate clusters when the annotation forest contains more than one partition tree.
According to our data model \eqref{scamp:cluster_distribution}, 
an observation belonging to multiple candidate clusters would indicate it is a sample from multiple different components of the mixture.
Since we do not allow this, SCAMP is forced to make a selection.

To make this selection, we define a preference function and use it to score each candidate cluster.
This is an opportunity for prior knowledge about the mixture distribution \eqref{scamp:cluster_distribution} to enter the clustering procedure:
the score reflects the user's belief about how the mixture is structured.
SCAMP's default preference function prizes candidate clusters with low-variance, symmetric coordinate distributions.
SCAMP uses a combination of the dip test and the sample L-moments of \cite{hosking1990moments,hosking2007some} to quantify this preference.
If a user is analyzing data where they expect the mixture to have a different composition, the preference function can be modified to better suit specific domains. 
For example, the preference function might be modified so that candidate clusters with skew distributions are preferred.

Once the candidate clusters are scored, selection is cast as an instance of the maximum-weight independent set problem.
SCAMP attempts to pick a collection of candidate clusters with maximal total preference score, subject to the constraint that no two selected clusters contain a common observation of the data matrix.
SCAMP settles for a greedy solution to this problem (preferring an approximate-but-quick solution to refinements like those discussed in \cite{sanghavi2008message} and \cite{brendel2010segmentation}).
This results in determining a subset of \textit{selected clusters} from the set of \textit{candidate clusters}.

This selection procedure often results in a large collection of residual observations that are not elements of any selected cluster.
From the perspective of the underlying data model, the residual observations are thought to be samples from different components of the mixture \eqref{scamp:cluster_distribution}.
To cluster these residual observations, the SCAMP search is repeated on the corresponding residual sub-matrix. 
This causes a new set of candidate clusters to be found, scored, and selected. 
These iterations continue until each row of the initial data matrix is an element of a unique selected cluster.

The final stage of the SCAMP procedure is annotation.
Since the default preference function causes SCAMP to prefer clusters with large dip-test p-values,
SCAMP assumes every selected $\alpha$-$m$-cluster is unimodal along each of its coordinates.
SCAMP annotates a selected cluster by comparing the median value along each of its coordinates with a collection annotation boundaries for the same coordinate (see section \ref{section:scamp} for further details). 

When the scale of measurement of a coordinate is meaningful, the annotation procedure can create clusters with interpretable labels.
Even when the coordinates are not directly interpretable (as is the case when SCAMP is applied to principal components), this annotation stage has the effect of merging clusters.
This merging can be justified in view of the data model: merging two selected clusters indicates that they were two independent samples from the same component of the mixture \eqref{scamp:cluster_distribution}.
The clustering returned by a single SCAMP iteration is made up of a disjoint collection of labeled clusters.

\subsection{Outline of Paper}

The remainder of this paper is organized as follows.
In section \ref{section:doubledip}, we develop an extended version of the dip test of \cite{hartigan1985dip} that is used by SCAMP to search for candidate clusters containing a small number of observations.
Section \ref{section:scamp} provides the details of the SCAMP clustering algorithm.
The performance of SCAMP is examined in section \ref{section:casestudes}, where we apply it to simulated and real data sets.
Proofs of several claims made in section \ref{section:doubledip} are contained in the appendix.

As mentioned earlier, SCAMP often uses the taut string of \cite{davies2001local,davies2004densities} to induce modal groups.
We decided to use the taut string since, as noted by \cite{davies2004densities} page 1099, observation (iii), it has the minimum modality over a large collection of density estimators.
We found this modal sparsity appealing given SCAMP's search strategy.
However, SCAMP always uses the default value of $\kappa=19$ (discussed and set by the authors in \cite{davies2004densities}) to solve the $\kappa$-Kuiper problem.

For data matrices with fewer than $400$ observations, 
the splitting is done with an extended version of the dip test in a sequential procedure designed to estimate the number of modes $k$ present in the coordinate of the sub-matrix under consideration. 
In part, our extension of the dip test grew out of a desire to minimize SCAMP's dependence on bandwidth selection.
Once estimated, we use $\hat{k}$ in conjunction with the one-dimensional dynamic programming version of $k$-medoids by \cite{wang2011ckmeans}
to induce modal sub-collections (and so ensure the exactly $\hat{k}$ modal sub-collections are produced). 

\begin{figure}[tbh]
\centering 
\begin{tikzpicture}[x=1pt,y=1pt]
\definecolor{fillColor}{RGB}{255,255,255}
\path[use as bounding box,fill=fillColor,fill opacity=0.00] (0,0) rectangle (361.35,216.81);
\begin{scope}
\path[clip] (  0.00,  0.00) rectangle (361.35,216.81);
\definecolor{drawColor}{RGB}{255,255,255}
\definecolor{fillColor}{RGB}{255,255,255}

\path[draw=drawColor,line width= 0.6pt,line join=round,line cap=round,fill=fillColor] (  0.00,  0.00) rectangle (361.35,216.81);
\end{scope}
\begin{scope}
\path[clip] ( 41.67, 29.59) rectangle (355.85,211.31);
\definecolor{fillColor}{RGB}{255,255,255}

\path[fill=fillColor] ( 41.67, 29.59) rectangle (355.85,211.31);
\definecolor{drawColor}{RGB}{0,0,0}

\path[draw=drawColor,line width= 2.8pt,line join=round] ( 55.95, 37.85) --
	( 57.85, 37.86) --
	( 59.76, 37.88) --
	( 61.66, 37.90) --
	( 63.56, 37.93) --
	( 65.47, 37.98) --
	( 67.37, 38.05) --
	( 69.28, 38.14) --
	( 71.18, 38.26) --
	( 73.08, 38.42) --
	( 74.99, 38.63) --
	( 76.89, 38.91) --
	( 78.80, 39.27) --
	( 80.70, 39.72) --
	( 82.61, 40.30) --
	( 84.51, 41.02) --
	( 86.41, 41.91) --
	( 88.32, 42.99) --
	( 90.22, 44.30) --
	( 92.13, 45.85) --
	( 94.03, 47.67) --
	( 95.93, 49.80) --
	( 97.84, 52.23) --
	( 99.74, 54.98) --
	(101.65, 58.06) --
	(103.55, 61.46) --
	(105.45, 65.14) --
	(107.36, 69.09) --
	(109.26, 73.26) --
	(111.17, 77.57) --
	(113.07, 81.98) --
	(114.98, 86.38) --
	(116.88, 90.68) --
	(118.78, 94.80) --
	(120.69, 98.63) --
	(122.59,102.07) --
	(124.50,105.03) --
	(126.40,107.43) --
	(128.30,109.20) --
	(130.21,110.29) --
	(132.11,110.67) --
	(134.02,110.34) --
	(135.92,109.30) --
	(137.83,107.59) --
	(139.73,105.27) --
	(141.63,102.42) --
	(143.54, 99.13) --
	(145.44, 95.51) --
	(147.35, 91.66) --
	(149.25, 87.72) --
	(151.15, 83.79) --
	(153.06, 80.02) --
	(154.96, 76.51) --
	(156.87, 73.37) --
	(158.77, 70.72) --
	(160.68, 68.65) --
	(162.58, 67.26) --
	(164.48, 66.61) --
	(166.39, 66.79) --
	(168.29, 67.85) --
	(170.20, 69.84) --
	(172.10, 72.79) --
	(174.00, 76.71) --
	(175.91, 81.61) --
	(177.81, 87.45) --
	(179.72, 94.20) --
	(181.62,101.78) --
	(183.52,110.09) --
	(185.43,119.01) --
	(187.33,128.38) --
	(189.24,138.02) --
	(191.14,147.72) --
	(193.05,157.29) --
	(194.95,166.47) --
	(196.85,175.05) --
	(198.76,182.80) --
	(200.66,189.52) --
	(202.57,195.03) --
	(204.47,199.18) --
	(206.37,201.87) --
	(208.28,203.05) --
	(210.18,202.71) --
	(212.09,200.91) --
	(213.99,197.74) --
	(215.90,193.36) --
	(217.80,187.96) --
	(219.70,181.78) --
	(221.61,175.07) --
	(223.51,168.09) --
	(225.42,161.11) --
	(227.32,154.41) --
	(229.22,148.22) --
	(231.13,142.76) --
	(233.03,138.21) --
	(234.94,134.71) --
	(236.84,132.35) --
	(238.75,131.17) --
	(240.65,131.16) --
	(242.55,132.24) --
	(244.46,134.33) --
	(246.36,137.25) --
	(248.27,140.85) --
	(250.17,144.89) --
	(252.07,149.16) --
	(253.98,153.42) --
	(255.88,157.44) --
	(257.79,161.00) --
	(259.69,163.89) --
	(261.59,165.94) --
	(263.50,167.02) --
	(265.40,167.03) --
	(267.31,165.92) --
	(269.21,163.67) --
	(271.12,160.31) --
	(273.02,155.92) --
	(274.92,150.60) --
	(276.83,144.48) --
	(278.73,137.71) --
	(280.64,130.45) --
	(282.54,122.87) --
	(284.44,115.14) --
	(286.35,107.42) --
	(288.25, 99.85) --
	(290.16, 92.56) --
	(292.06, 85.64) --
	(293.97, 79.19) --
	(295.87, 73.25) --
	(297.77, 67.86) --
	(299.68, 63.03) --
	(301.58, 58.78) --
	(303.49, 55.06) --
	(305.39, 51.87) --
	(307.29, 49.15) --
	(309.20, 46.87) --
	(311.10, 44.97) --
	(313.01, 43.42) --
	(314.91, 42.16) --
	(316.82, 41.15) --
	(318.72, 40.35) --
	(320.62, 39.72) --
	(322.53, 39.24) --
	(324.43, 38.87) --
	(326.34, 38.58) --
	(328.24, 38.37) --
	(330.14, 38.22) --
	(332.05, 38.10) --
	(333.95, 38.02) --
	(335.86, 37.96) --
	(337.76, 37.92) --
	(339.66, 37.89) --
	(341.57, 37.87);

\path[draw=drawColor,line width= 2.8pt,line join=round] (185.43,119.01) -- (283.49,119.02);
\definecolor{drawColor}{gray}{0.20}

\path[draw=drawColor,line width= 0.6pt,line join=round,line cap=round] ( 41.67, 29.59) rectangle (355.85,211.31);
\end{scope}
\begin{scope}
\path[clip] (  0.00,  0.00) rectangle (361.35,216.81);
\definecolor{drawColor}{gray}{0.30}

\node[text=drawColor,anchor=base east,inner sep=0pt, outer sep=0pt, scale=  0.88] at ( 36.72, 34.79) {0.00};

\node[text=drawColor,anchor=base east,inner sep=0pt, outer sep=0pt, scale=  0.88] at ( 36.72, 80.41) {0.05};

\node[text=drawColor,anchor=base east,inner sep=0pt, outer sep=0pt, scale=  0.88] at ( 36.72,126.03) {0.10};

\node[text=drawColor,anchor=base east,inner sep=0pt, outer sep=0pt, scale=  0.88] at ( 36.72,171.64) {0.15};
\end{scope}
\begin{scope}
\path[clip] (  0.00,  0.00) rectangle (361.35,216.81);
\definecolor{drawColor}{gray}{0.20}

\path[draw=drawColor,line width= 0.6pt,line join=round] ( 38.92, 37.82) --
	( 41.67, 37.82);

\path[draw=drawColor,line width= 0.6pt,line join=round] ( 38.92, 83.44) --
	( 41.67, 83.44);

\path[draw=drawColor,line width= 0.6pt,line join=round] ( 38.92,129.06) --
	( 41.67,129.06);

\path[draw=drawColor,line width= 0.6pt,line join=round] ( 38.92,174.67) --
	( 41.67,174.67);
\end{scope}
\begin{scope}
\path[clip] (  0.00,  0.00) rectangle (361.35,216.81);
\definecolor{drawColor}{gray}{0.20}

\path[draw=drawColor,line width= 0.6pt,line join=round] ( 55.95, 26.84) --
	( 55.95, 29.59);

\path[draw=drawColor,line width= 0.6pt,line join=round] (151.15, 26.84) --
	(151.15, 29.59);

\path[draw=drawColor,line width= 0.6pt,line join=round] (246.36, 26.84) --
	(246.36, 29.59);

\path[draw=drawColor,line width= 0.6pt,line join=round] (341.57, 26.84) --
	(341.57, 29.59);
\end{scope}
\begin{scope}
\path[clip] (  0.00,  0.00) rectangle (361.35,216.81);
\definecolor{drawColor}{gray}{0.30}

\node[text=drawColor,anchor=base,inner sep=0pt, outer sep=0pt, scale=  0.88] at ( 55.95, 18.58) {-10};

\node[text=drawColor,anchor=base,inner sep=0pt, outer sep=0pt, scale=  0.88] at (151.15, 18.58) {-5};

\node[text=drawColor,anchor=base,inner sep=0pt, outer sep=0pt, scale=  0.88] at (246.36, 18.58) {0};

\node[text=drawColor,anchor=base,inner sep=0pt, outer sep=0pt, scale=  0.88] at (341.57, 18.58) {5};
\end{scope}
\begin{scope}
\path[clip] (  0.00,  0.00) rectangle (361.35,216.81);
\definecolor{drawColor}{RGB}{0,0,0}

\node[text=drawColor,anchor=base,inner sep=0pt, outer sep=0pt, scale=  1.10] at (198.76,  5.50) {x};
\end{scope}
\begin{scope}
\path[clip] (  0.00,  0.00) rectangle (361.35,216.81);
\definecolor{drawColor}{RGB}{0,0,0}

\node[text=drawColor,rotate= 90.00,anchor=base,inner sep=0pt, outer sep=0pt, scale=  1.10] at ( 13.08,120.45) {f(x) };
\end{scope}
\end{tikzpicture}
\caption[Excess mass will not detect multimodality]{As observed in \cite{hall2004attributing}, the excess mass difference $\Delta_3=0$ iff the smallest mode of the distribution lies below the base of the valley of the larger two modes.
  In the depicted Gaussian mixture, the excess mass difference does not detect trimodality since $E_3(\lambda)=E_2(\lambda)$ for values of $\lambda$, including those above the horizontal line. }
\label{fig:excessmassfail} 
\end{figure}
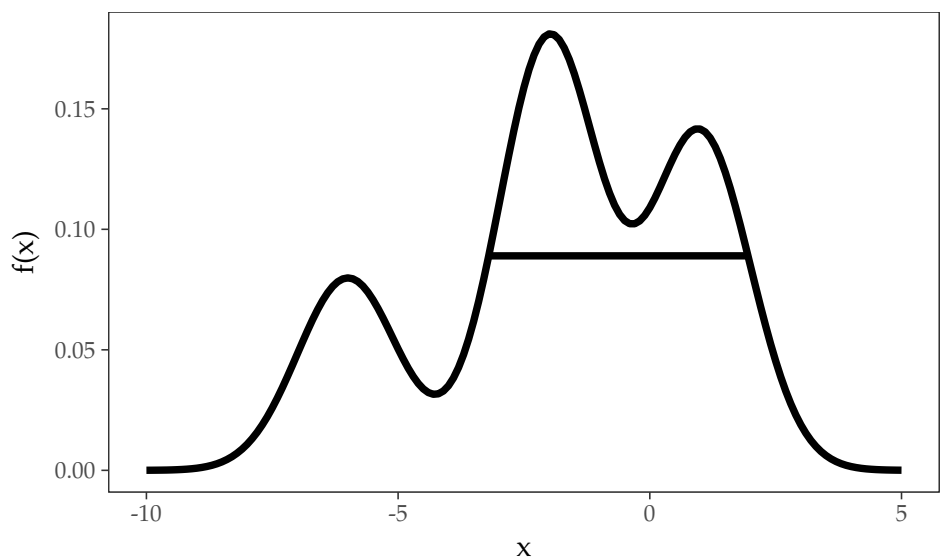

To our knowledge, this extended version of the dip test is new.
Other testing procedures, such as the excess mass test, could have been used in its place:
\cite{ameijeiras2016mode} discuss additional testing procedures.
However, as noted by \cite{hartigan2000testing} as well as \cite{hall2004attributing}, the excess mass test concludes the presence of two modes in situations like those depicted by figure \ref{fig:excessmassfail}.
SCAMP is designed with the goal of detecting such small groups and selecting them if they have high preference scores: using our extension makes this outcome possible.
We also considered using the multiscale procedure of 
\cite{dumbgen2008multiscale}
and the jaws-based method of 
\cite{hartigan2000testing} to test for multimodality in small samples.
We decided against using these tests since they both required developing automated procedures to make decisions on the basis of interval collections:
in the former, the collections $\mathcal{D}^{+}(\alpha)$ and $\mathcal{D}^{-}(\alpha)$;
in the latter, $W$- an $M$-components contained within larger candidate antimodal sections of the empirical distribution.
Thinking about how to develop such automated procedures led us to consider our extension to the dip, which we discuss in the next section.

\subsection{Summary of Contributions}

In this paper, we have developed a new clustering algorithm called selective clustering annotation using modes of projections (SCAMP). SCAMP is designed to be:

\begin{itemize}
  \item{\textbf{Interpretable.} Clusters produced by SCAMP are automatically assigned labels in terms of the columns of the underlying dataset.
      When the measurements in the underlying dataset are themselves meaningful, this produces clusters that human users can understand simply by reading their label.}
  \item{\textbf{Robust.} SCAMP defines clusters in terms of the empirical distributions of their coordinate projections: they must be unimodal.
      SCAMP is consequently able to produce clusters with long tails that accommodate outliers.}
  \item{\textbf{Heterogeneous.} SCAMP is able to produce clusterings of data matrix consisting of clusters of different sizes.
      }
  \item{\textbf{Simple to use.} SCAMP determines the number of clusters in a dataset  as a consequence of three tuning parameters: a level $\alpha$, 
      against which p-values from dip test are compared; a lower-bound $m$,
      which defines the smallest number of observations that can constitute a cluster; and a 
      univariate Gaussian variance parameter $\gamma$.
      The parameter $\alpha$ describes a univariate quantity, no matter the dimension 
      of the data matrix.
      Given these parameters, SCAMP will cluster the dataset:  
      estimating the number of clusters in a data matrix is an implicit part of its clustering 
      strategy.
      This is in contrast to procedures such as $k$-means and $k$-medoids, 
      which either require the user to set the number of clusters $k$ themselves, or to estimate 
      this parameter prior to clustering.}
  \item{\textbf{Composable.} The most computationally demanding task in the SCAMP procedure is the 
  search for candidate clusters.
      Each tree grown in the search requires computing the dip test and taut string density 
      estimator numerous times. 
      Fortunately, $O(n)$ algorithms exist for both computations.
      In addition, the search can be parallelized: a parallel \CC \  implementation of this 
      procedure with R interface is available.
      When combined with a random-sampling scheme, it is possible to use SCAMP to cluster datasets 
      with a large number of observations
      and moderate number of features. We demonstrate this possibility on both real and simulated 
      data in section \ref{section:casestudes}.
      In modern datasets, where the number of features is much larger than the number of 
      observations, 
      SCAMP can be used productively 
      in conjunction with dimensionality reduction methods such as 
      PCA.
      We demonstrate this possibility on bulk RNA seq data in section \ref{section:casestudes}.}
\end{itemize}

\section{Double Dipping}\label{section:doubledip}

To introduce our extension of the dip test, we start with an informal discussion of the dip itself.
The basic observation underpinning the dip test of \cite{hartigan1985dip} is that the distribution function (df)
of a continuous unimodal distribution (suitably defined) has an s-shape.
Suppose a unimodal df with support on $\mathbb{R}$ has a unique mode $m$.
On the interval $(-\infty,m]$  such a df is convex; on $[m,\infty)$, it is concave. Taken together, they form the s-shape. 

Independent samples from a unimodal df will produce empirical dfs that resemble the underlying s-shape with increasing fidelity as the sample size grows larger.
The dip takes advantage of the fact that the greatest convex minorant (GCM) and least concave majorant (LCM) of an empirical distribution function:
\begin{enumerate}
  \item{Are inexpensive to compute.}
  \item{Do not require the user to select any tuning parameters in order to compute them.}
  \item{Can be combined with a linear component in order to estimate a sample's departure from unimodality under the null hypothesis that they originate from a unimodal (s-shaped) df.}
\end{enumerate}

\begin{figure}[tbh]
\centering 
\input{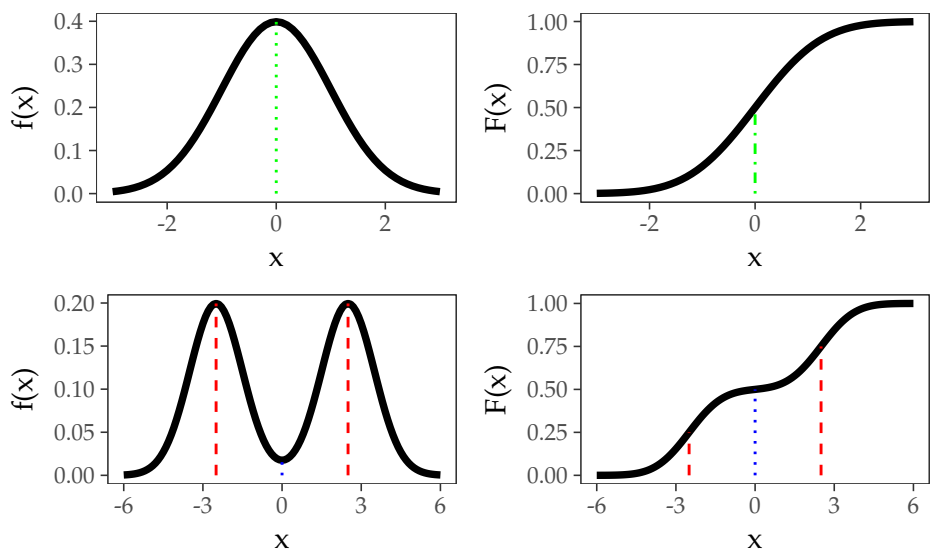}
\caption[Unimodal and Bimodal intuition]{The unimodal density and df identifies its unique mode $m$ with the green dotted-dashed line. The bimodal density and df identify modes $m_1<m_2$ with dashed red lines, and
  the antimode $c_1$ with a blue dotted line. Compare the $s$-shape of the unimodal distribution the double-$s$ of the bimodal distribution function, with the antimode marking the split in the bimodal case.}
\label{fig:doubledip} 
\end{figure}

Now, suppose we are examining a bimodal df $F$ with support on $\mathbb{R}$ that has two unique modes $m_1,\ m_2,$ and antimode $c_1$, such that $m_1 < c_1 < m_2$.
The idea behind our extension is the observation that such a distribution function consists of two unimodal sub-dfs stitched together:
on the interval $(-\infty,m_1]$ it is convex; on $[m_1,c_1]$, concave; on $[c_1,m_2]$ it is again convex; and concave again on $[m_2,\infty)$.
Restricting to the interval $(-\infty,c_1]$ yields one $s$-shaped sub-df; restricting to $[c_1,\infty)$ another.
See figure \ref{fig:doubledip} for an example.

Continuing with this bimodal $F$, we define
\begin{align}
 H_1 \equiv  H_{(-\infty,c_1]}= \frac{F}{F(c_1)}\cdot 1_{(-\infty,c_1)}+1_{[c_1,\infty)} \label{section:doubledip:r1}
\end{align}
and
\begin{align}
  H_2 \equiv H_{[c_1,\infty)}=\frac{F-F(c_1)}{1-F(c_1)}\cdot 1_{[c_1,\infty)} \ .\label{section:doubledip:r2}
\end{align}
Both $H_1$ and $H_2$ are unimodal dfs in their own right.
From this it follows that the $1$-dip will tend to $0$ if applied to data from either $H_{(-\infty,c_1]}$ or $H_{(c_1,\infty)}$.
This suggests the $1$-dip algorithm of \cite{hartigan1985dip} might be used to compute the $2$-dip as well: we expect 
the maximum estimated dip of samples from the restrictions $H_{(-\infty,c_1]}$ and $H_{(c_1,\infty)}$ to be small.
Notice that this observation extends to an arbitrary (fixed) number of modes: any multimodal distribution (suitably defined) can be split at its antimodes into unimodal component dfs (when scaled).
Moreover, the dip can be applied to each of these restrictions in turn.
This suggests the following extension of the dip test.
Under the hypothesis of data being generated from a distribution with $k$-modes, split a sorted sample into $k$ sub-collections at its $k-1$ antimodes and apply the $1$-dip to each sub-collection.
If the $k$-mode hypothesis is true, we expect each of these restricted dips to be small. 

The remainder of this section extends the theorems of 
\cite{hartigan1985dip} to justify the preceding informal development. We extend the theorems under the assumption the location of antimodes are known.
Under such an assumption, the theorems of \cite{hartigan1985dip} extend naturally: we essentially inherit the original arguments.
In fact, most of the work is notational, in order to accommodate a distribution with multiple modes.
We develop necessary notation for the remainder of this section (taking the notation directly from \cite{hartigan1985dip}),
and conclude with a heuristic development of the algorithm we use to estimate the $N$-dip in practice.

For bounded functions $F, G$ on $\mathbb{R}$ define $\rho(F,G) \equiv \sup_x |F(x)-G(x)|$. For a collection of bounded functions $\mathcal{A}$, define
\[
  \rho(F,\mathcal{A}) \equiv \inf_{A \in \mathcal{A}} \rho(F,A)\ .
\]
Two classes have special importance.

\begin{defn}[The class $\mathcal{U}_N$]\label{uclass}
  For $N \in \mathbb{N}$, let $\mathcal{U}_N$ be the set of all distribution functions $F$ on $\mathbb{R}$
  where there exist real numbers
  \[
    m_1 \leq c_1 \leq m_2 \leq \ldots \leq m_{N-1} \leq c_{N-1} \leq m_N
  \]
  such that $F$ is convex on $(-\infty, m_1]$, concave on $[m_N, \infty)$, convex on $[c_{j-1},m_j]$ with
  $2 \leq j \leq N$, and concave on $[m_j,c_j]$ with $1 \leq j \leq N-1$.
\end{defn}

\begin{defn}[The class $\mathcal{V_N}$]\label{vclass}
  For $N \in \mathbb{N}$, let $\mathcal{V}_N$ be the set of all functions $f$ on $\mathbb{R}$
  where there exist real numbers
  \[
    0\equiv c_0 \leq  m_1 \leq c_1 \leq m_2 \leq \ldots \leq m_{N-1} \leq c_{N-1} \leq m_N \leq c_N\equiv 1
  \]
  such that $f$ is a finite constant on $(-\infty, 0]$, (a possibly different) constant on $[1, \infty)$,
  non-decreasing on $(0,1)$,
  convex on $[c_{j-1},m_j]$ with $2 \leq j \leq N$ and concave on $[m_j,c_j]$ with $1 \leq j \leq N-1$.
\end{defn}

We will extend the dip relative to the first class $\mathcal{U}_N$.

\begin{defn}[$N$-dip]\label{Ndip}
  For fixed $N \in \mathbb{N}$, define the $N$-dip of a distribution function $F$ to be
  \[
    D_N(F) \equiv \rho(F,\mathcal{U}_N)\ .
  \]
\end{defn}

Note that when $N=1$, the preceding definitions reduce to those in the paper of \cite{hartigan1985dip}.
Also note the classes are nested: for all $N \in \mathbb{N}$, 
$\mathcal{U}_N \subset \mathcal{U}_{N+1}$
and $\mathcal{V}_N \subset \mathcal{V}_{N+1}$.
For the remainder of this section, suppose $N \geq 2$, $N\in \mathbb{N}$, and that $N$ is fixed.
The observations made in \cite{hartigan1985dip} persist for the $N$-dip: for distributions $F_1,F_2$ we have
\[
  D_N(F_1)
  = \inf_{G \in\ \mathcal{U_N}} \sup_x |F_1-F_2+F_2-G| 
  \leq D_N(F_2) + \rho(F_1,F_2)\ .
\]
Moreover, for $F \in \mathcal{U}_N$ we have $D_N(F) = 0$, so $D_N(F)$ measures the departure of $F$ from distributions with $N$ modes.
In what follows, let $\mathcal{C}^{+}_{[a,b]}$ denote the collection of all convex functions on the interval $[a,b]$, and
$\mathcal{C}^{-}_{[a,b]}$ the concave functions on the interval $[a,b]$, with $a,b \in \mathbb{R}$, $a<b$.
If either $a= -\infty$ or $b=\infty$, open the interval to extend the notation. We will use the GCM and LCM to measure departures from a modal hypothesis.

\begin{defn}[Greatest Convex Minorant (GCM)]
  The greatest convex minorant (GCM) of a function $f$ on an interval $[a,b]$ is defined
  \[
    GCM(F,[a,b]) \equiv \sup\left\{G(x)\ |\ G(x) \leq F(x) \text{ on } [a,b], \text{ and } G(x) \in \mathcal{C}^{+}_{[a,b]}\right\}\ .
  \]
\end{defn}

\begin{defn}[Least Concave Majorant (LCM)]
  The least concave majorant (LCM) of a function $f$ on an interval $[a,b]$ is defined
  \[
    LCM(F,[a,b]) \equiv \inf\left\{G(x)\ |\ G(x) \geq F(x) \text{ on } [a,b], \text{ and } G(x) \in \mathcal{C}^{-}_{[a,b]}\right\}\ .
  \]
\end{defn}

With these definitions in place, we first state a bound on the $N$-dip.

\begin{lem}\label{lem:dip_bound}
  For $N \geq 2$, $N \in \mathbb{N}$ we have for any distribution function $F$ on $\mathbb{R}$, we have
  \[
    D_N(F) \leq \frac{1}{2N}\ .
  \]
\end{lem}

Proof of Lemma \ref{lem:dip_bound}, along with the remaining claims in this section, are given in Appendix \ref{section:appendix}.
The proofs themselves are natural extensions of those given by \cite{hartigan1985dip}.
Our development and presentation of these extensions were also shaped by unpublished course notes of \cite{dudleynotes2015},
available on his website.

We next observe for distributions concentrated on $[0,1]$, departures from
the classes $\mathcal{U}_N$ and $\mathcal{V}_N$ are equivalent.

\begin{lem}\label{lem:d_dd_equiv}
  Let $N \in \mathbb{N}$ be fixed. Let $F$ be a distribution function with $F(0)=0$ and $F(1)=1$. Then
  \[
    \rho(F,\mathcal{U}_N) = \rho(F,\mathcal{V}_N)\ .
  \]
\end{lem}

As with the dip, we next see that scaling and mixing the Uniform distribution with suitably bounded function scales the $N$-dip.

\begin{lem}\label{lem:uniform_mix}
  Let $F$ be a bounded function constant on $[-\infty,0]$ and on $[1,\infty]$. Let $I$ be the distribution
  function of the Uniform on $(0,1)$. Let $N \in \mathbb{N}$. Then
  \[
    D_N(\alpha F + \beta I) = \alpha D_N(F)
  \]
  for $\alpha, \beta \geq 0$. 
\end{lem}

The first theorem states that, as with the dip, the Uniform distribution is also extreme for the $N$-dip.

\begin{thm}\label{thm:conv_bb}
  Let $N \in \mathbb{N}$.
  Let $F_n$ be the empirical distribution function for a sample of size $n$ from the Uniform on $(0,1)$, and let $B$ be
  the Brownian Bridge process with $\text{cov}[B(s),B(t)] = s(1-t)$ for $0 \leq s \leq t \leq 1$. Assume that $B$ is zero
  outside $(0,1)$.  Then
  \[
    \sqrt{n}D_N(F_n) \rightarrow_d D_N(B) \text{ as } n \nearrow \infty\ .
  \]
\end{thm}

The next theorem states that the $N$-dip, when applied to samples from multimodal distributions with exponential changes between modes and exponential decrease away from extreme modes, will tend to zero in probability.

\begin{thm}\label{thm:dip_to_zero}
  Let $N > 2, N \in \mathbb{N}$ be fixed. Let $F \in \mathcal{U}_N$, and have non-zero
  $k^{th}$ derivative at the distinct antimodes $c_1,\ldots,c_{N-1}$ and the
  distinct modes $m_1,\ldots,m_N$, for some $k \geq 2$. For each $\epsilon, \epsilon_1,
  \epsilon_2 > 0$, and for $j \in \left\{1,2,\ldots,N\right\}$ that matches 
  the number of modes and antimodes, suppose
  \begin{align}
    \inf_{x < m_1-\epsilon}\ &\vline \frac{d}{dx} \log F'(x) \vline > 0 \ ,  \label{thm:dip_to_zero:c1} \\
    \inf_{x > m_N+\epsilon}\ &\vline \frac{d}{dx} \log F'(x) \vline > 0 \ ,  \label{thm:dip_to_zero:c2} \\
    \inf_{c_{j-1}+\epsilon_1 < x < m_j-\epsilon_2}\ &\vline \frac{d}{dx} \log F'(x) \vline > 0 \ , \text{ and } \label{thm:dip_to_zero:c3} \\
    \inf_{m_j+\epsilon_1 < x < c_{j+1}-\epsilon_2}\ &\vline \frac{d}{dx} \log F'(x) \vline > 0 \ . \label{thm:dip_to_zero:c4} 
  \end{align}
  Then $\sqrt{n}D_N(F_n) \rightarrow_p 0$.
\end{thm}

To summarize our findings, theorem \ref{thm:dip_to_zero} suggests the $N$-dip will be 
able to detect a large number of distributions asymptotically.
Theorem  \ref{thm:conv_bb} suggests the Uniform distribution may serve well as the null
distribution in finite samples. 
How do we compute the $N$-dip in finite samples? Theorem 6 of \cite{hartigan1985dip} 
and the subsequent discussion provide an algorithm for computing the $1$-dip
[\cite{hartigan1985algorithm,maechler2009diptest}].
Here we suggest an extension to estimate the $N$-dip. For notational clarity, we focus
our discussion on the $2$-dip.

Returning to the bimodal example that began the section, we again consider the
restrictions \eqref{section:doubledip:r1} and \eqref{section:doubledip:r2}.
Observe that
\[
F = H_{1}\cdot F(c_1) +H_{2}\cdot (1-F(c_1)) \ .
\]
If we knew the location of the antimode $c_1$, the empirical distribution could be similarly partitioned:
\[
F_n = H_{n,1}\cdot F(c_1) +H_{n,2}\cdot (1-F(c_1)) \ ,
\]
with $H_{n,1}$ and $H_{n,2}$ formed by $F_n$ replacing $F$ in the definitions of $H_1$ and $H_2$.
But, subject to the conditions of  theorem \ref{thm:dip_to_zero}, $\sqrt{n}D_1(H_{n,1}) \rightarrow_p 0$ and $\sqrt{n}D_1(H_{n,2}) \rightarrow_p 0$.
Combined with Lemma \ref{lem:uniform_mix}, we infer
\begin{align}
\sqrt{n}D_2(F_n) 
&= \sqrt{n}\cdot F(c_1)\cdot D_2(H_{n,1})  \vee \sqrt{n}\cdot (1-F(c_1))\cdot D_2(H_{n,2})  \nonumber \\
&= \sqrt{n}\cdot F(c_1)\cdot D_1(H_{n,1})  \vee \sqrt{n}\cdot (1-F(c_1))\cdot D_1(H_{n,2})  \rightarrow_p 0\ ,
\end{align}
where the second equality is suggested by the fact that
$\mathcal{V}_1 \subset \mathcal{V}_2$.
Note the  order statistics $X_{(1)},X_{(2)},\ldots,X_{(n)}$ are consistent for their population counterparts: as the sample size increases a sample point will grow arbitrarily close to $c_1$.
In finite samples, we lack knowledge of $F$ and $c_1$. Under the hypothesis of bimodality, plugging in empirical estimates leads naturally to 
\[
\sqrt{n}\hat{D}_2(F_n) = \sqrt{n}\cdot F_n(\hat{c}_1)\cdot D_1(H_{n,1})  \vee \sqrt{n}\cdot (1-F_n(\hat{c}_1))\cdot D_1(H_{n,2})\ .
\]
Since we do not make assumptions about the relative magnitude or location of the modes, the estimate $\hat{c}_1$ can be obtained by exhaustive data splitting.
Hence, for $j \in \left\{4,5,\ldots,n-3\right\}$,  let $\hat{c}_{1,j} = X_{(j)}$. 
The hypothesis of bimodality indicates that splitting the sample at the order statistic closest to the antimode will lead to scaled unimodal distributions that each minimize the maximum $1$-dip of $H_{n,1}$ and $H_{n,2}$.
Our proposed estimator of the $2$-dip is consequently
\[
\sqrt{n}\hat{D}_2(F_n) = \min_{j \in \left\{4,5,\ldots,n-3\right\}}\left[\sqrt{n}\cdot F_n(\hat{c}_{1,j})\cdot D_1(H_{n,1})  \vee \sqrt{n}\cdot (1-F_n(\hat{c}_{1,j}))\cdot D_1(H_{n,2})\right]\ .
\]
The restriction on $j$ follows because the $1$-dip achieves its lower bound with high probability on small sample sizes $4$ (see Proposition 1 of \cite{dudleynotes2015}).
Conveniently, this formulation allows us to use the $1$-dip algorithm of Hartigan and Hartigan to estimate the $2$-dip. 
The formulation also extends naturally to the estimation of the $N$-dip, by partitioning order statistics of an observed sample into $N$-components (with each component containing at least $4$ observations),
weighting them according to their empirical mass, and then searching for the sample index that minimizes the maximum over the collection.

For fixed sample size, theorem \ref{thm:conv_bb} suggests the Uniform can be used to obtain p-value.
We use the Uniform distribution in practice: we have computed the distribution of
values obtained by the estimator on random samples from the Uniform distribution for
sample sizes up to $N = 400$ for the $2$-dip and $3$-dip. 
These values, along with the implementation of the $2$-dip and $3$-dip estimators, are
available in the package \pkg{scamp}.
Pseudo-code describing the computation of the $2$-dip-estimate is given in figure \eqref{doubledip:doubleDip}. 

\begin{algorithm}\label{doubleDip}
  \caption{$2$-dip}\label{doubledip:doubleDip}
  \begin{algorithmic}[1]
    \Function{doubleDip}{$\mathbf{x}$}\Comment{$\mathbf{x}$ is a vector with no ties sorted in increasing order.}
    \State $\text{bestDip} \gets 1$
    \State $N \gets \text{length}(\mathbf{x})$
    \For{$i \text{ in } \left\{4,5,\ldots,N-4\right\}$}
      \State $\mathbf{x_1} \gets x\text{[}1\text{:}i\text{]}$
      \State $\mathbf{x_2} \gets x\text{[}i\text{:}N-4\text{]}$
      \State $\mathbf{d_1} \gets \left(\text{hartiganDip}(\mathbf{x_1})\right)\cdot \left(i/\sqrt{N}\right)$
      \State $\mathbf{d_2} \gets \left(\text{hartiganDip}(\mathbf{x_2})\right)\cdot \left([N-i]/\sqrt{N}\right)$
      \State $\mathbf{d} \gets d_1 \vee d_2$
      \If {$d < \text{bestDip}$}
      \State $\text{bestDip} \gets d$
      \EndIf
      \EndFor
      \State $\text{return(bestDip)}$
      \EndFunction
  \end{algorithmic}
\end{algorithm}

\section{Selective Clustering Annotated using Modes of Projections}\label{section:scamp} 

SCAMP is an iterative procedure: each iteration determines a labeled partitional clustering of a data matrix $X$.
To understand its iterative nature, we must first describe details of a single SCAMP iteration that were omitted in section \ref{section:introduction}.
We do so now.

\subsection{Single Iteration}

Before SCAMP is applied to a data matrix $X$, there is a pre-processing step: the data matrix is normalized so each column vector has mean zero and unit variance.
After normalization, the data matrix is passed to SCAMP.

\begin{figure}[H]
  \centering
  \includegraphics[width=0.95\textwidth,keepaspectratio]{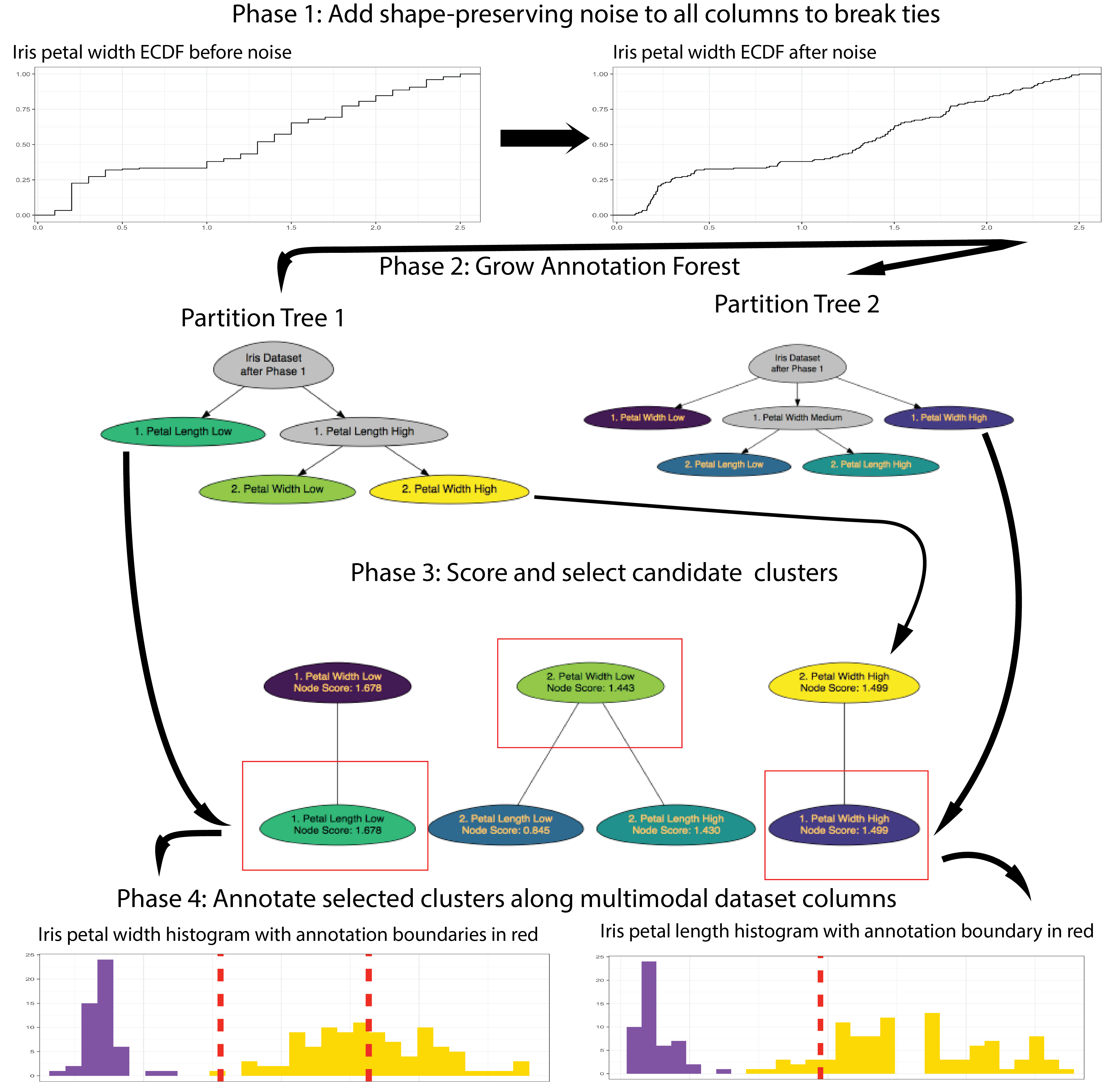}
  \caption[SCAMP outline]{A single SCAMP iteration applied the iris data set of \cite{fisher1936use}.
  The data set contains measurements of sepal length, sepal width, petal length, and petal width of three iris species.
  Measurements are taken to two decimal places, leading to multiple ties.
  The structured noise of phase 1 eliminates ties.
  The annotation forest depicted in phase 2 is exhaustive: all partition trees found by SCAMP are displayed.
  The candidate clusters are the leaves of the two partition trees.
  These candidate clusters are used to induce the graph of phase 3: an edge connects two candidate clusters if they share a common observation.
  The candidate clusters are then scored with a preference function.
  Preference scores are reported on each node.
  Selected clusters, bounded by red rectangles, are chosen to maximize the total sum of preference scores (candidate clusters with the same score are identical subsets of observations).
  In phase 4, the selected clusters are annotated by comparing them to annotation boundaries derived from the annotation forest.}\label{figure:scampOutline}
\end{figure}

There are four phases in a single SCAMP iteration. 
In the first phase, two forms of structured noise are added to the data matrix.
This is done so that ties within column vectors of a data matrix are broken and the relative order of observations within a column are perturbed.
In the second phase, the data matrix is recursively searched for $\alpha$-$m$-clusters.
In the third phase, a partial clustering of the data matrix is determined by selecting $\alpha$-$m$-clusters with high preference scores.
If any observations are not assigned to a cluster after phase three,
the second and third phase repeat on residual sub-matrices until a complete partitional clustering of the data matrix has been found.
In the fourth and final phase, the selected clusters are assigned descriptive labels.
This labeling provides a name for every row of the data matrix.
Figure \ref{figure:scampOutline} visualizes these four phases, when SCAMP is applied to the famous iris data set of \cite{fisher1936use}. 

We now look at the details of each phase.

\subsubsection{Add noise to the data matrix}\label{scamp:noiseStep}

For $1 \leq i \leq k$, denote the rows of a data matrix $X_{k \times p}$ by 
\[
r_i \equiv (x_{(i,1)},\ldots,x_{(i,k)})\ ,
\]
for $1 \leq j \leq p$, the columns by
\[
c_j \equiv (x_{(1,j)},\ldots,x_{(k,j)})' \ ,
\]
and specific entries as $x_{(i,j)}$. 
A single scamp iteration begins by adding noise to each column $c_j$ of the data matrix.
The noise procedure is the same for each column $c_j$. 
First, the order statistics of $c_j$ 
\[
O_{(1:d_1)}, O_{(2:d_2)},\ldots,O_{(q:d_q)}
\]
are determined. 
Here $O_{(1:d_1)}$ indicates the minimum value observed in column $c_j$ with $d_1$ repetitions, $O_{(q:d_q)}$ the maximum with $d_q$ repetitions, and $O_{(n:d_n)}$ 
for $1 < n < q$  the values between. Once determined, duplicate entries of $c_j$ are replaced by samples from a uniform distribution.
Suppose an entry $x_{(i,j)}$ corresponds to the order statistic $O_{(n:d_n)}$. If $d_n=1$, $x_{(i,j)}$ is unmodified. Otherwise $x_{(i,j)}$ is replaced by
\begin{align}
u_{(i,j)} \sim \text{Uniform}\left(\frac{O_{(n:d_n)}+O_{(n-1:d_{n-1})}}{2},\frac{O_{(n:d_n)}+O_{(n+1:d_{n+1})}}{2}\right)\ . \label{scamp:uniformNoise}
\end{align}
Notice that $d_n$ entries of $c_j$ are replaced by samples from this specific uniform distribution.
In the case that either the minimum value of $c_j$ is not unique, the lower bound of \eqref{scamp:uniformNoise} is replaced by that minimum value.
The same hold if the maximum value of $c_j$ has copies, though with the upper bound of \eqref{scamp:uniformNoise} replaced by the maximum value instead.

The choice of adding uniform noise is motivated by SCAMP's reliance on the dip test. 
As discussed in section \ref{section:doubledip}, the dip test is calibrated against the uniform distribution.
By replacing tied observations with draws from a uniform distribution that is supported on an interval bounded by neighboring order statistics, 
we have replaced a jump of $d_n/k$ in $c_j$'s empirical distribution with $d_n$ jumps of $1/k$. 
From the point of view of the dip test, the empirical distribution of $c_j$ now behaves like a unimodal distribution on the interval 
$([O_{(n:d_n)}+O_{(n-1:d_{n-1})}]/2,[O_{(n:d_n)}+O_{(n+1:d_{n+1})}]/2)$. 

Once uniform noise has been added to $c_j$, the order statistics 
\[
N_{(1:1)}, N_{(2:1)},\ldots,N_{(k:1)}
\]
are again determined. Notice each order statistic is now unique. 
The noise-step concludes by associating each entry 
$u_{(i,j)}$ with its new order statistic $N_{(n:1)}$, and replacing it by
\begin{align}
n_{(i,j)} \sim \text{Normal}\left(\mu = u_{(m,j)} , \sigma = \frac{N_{n+1:1}+N_{n-1:1}}{2\cdot \gamma}\right)\ , \label{scamp:GaussianNoise}
\end{align}
with $\gamma$ a tuning parameter (by default we set $\gamma \equiv 4$).
As almost all the mass is concentrated within three standard deviations of the mean, the addition of Gaussian noise \eqref{scamp:GaussianNoise} provides
observations with the chance to swap relative position with some of their neighbors. 
For observations with values near modes of $c_j$, this step is largely inconsequential in their ultimate cluster assignment. 
But for observations near antimodes of $c_j$, this step increases the chance they change cluster assignment on successive SCAMP iterations.
Over multiple SCAMP iterations, changes in cluster assignments provide an uncertainty measure of each observation's SCAMP label.

\subsubsection{Search for candidate clusters}\label{scamp:phase2}

After structured noise has been added to the data matrix, SCAMP searches the data matrix for subsets of rows that follow the definition of an $\alpha$-$m$-cluster.
To conduct this search, two parameters must be set: the significance level for the dip test of  \cite{hartigan1985dip}, $\alpha$, and the lower bound on cluster size, $m$. 

Once set, SCAMP begins by testing each coordinate projection of the dataset for multimodality using the dip test.
For each column in the data matrix, when multimodality is detected by the dip test
modal groups are induced by estimating the density of the column vector and then separating observations into groups relative to the antimodes of the density.
This is done using the taut string density estimator of \cite{davies2004densities}.
The cut-points that separate the modal groups are determined by calculating the average coordinate value along each antimodal component of the taut string.
Sub-matrices are then created for all observations that fall between each neighboring pair of cut-points (the minimum and maximum value along the coordinate sub-collection are treated as cut-points).
Within each sub-matrix, projections of the remaining $p-1$ coordinates are recursively tested for multimodality (again using the dip test at level $\alpha$).

If multimodality is detected in a coordinate of a sub-matrix containing fewer than $400$ observations, the taut-string is not used to induce modal groups.
Instead, bimodality is tested using the $2$-dip of section \ref{section:doubledip} at level-$\alpha/2$.
Failure to reject the hypothesis of bimodality causes SCAMP to conclude $\hat{k}=2$ modal sub-collections are present in the coordinate.
If bimodality is rejected, trimodality is tested using the $3$-dip at level-$\alpha/3$.
Now, a failure to reject leads SCAMP to conclude $\hat{k}=3$ modal sub-collections are present.
In principle, this sequential testing procedure can be carried out until it arrives at an estimate of $n$-modality.
However if tri-modality is rejected, SCAMP simply concludes $\hat{k}=4$ modal sub-matrices are present due to computational limitations.
SCAMP then separates the projection into modal sub-collections using
the one-dimensional version of $k$-medoids by \cite{wang2011ckmeans} with $\hat{k}$ groups.

We introduce this secondary sequential testing procedure in sub-matrices with fewer than $400$ observations for several reasons.  The first is computational. As the number of observations increases, the cost of exhaustive splitting grows with the falling-factorial of the number of modes of the $N$-dip. Because of this computational cost, we have not yet been able to compute tables of critical values for the $N$-dip when $N > 3$. 

Ideally, we would like to apply this sequential procedure until the $N$-dip fails to reject its null hypothesis for some $N$, in place of the current combination of the dip test and subsequent density estimate. This is because the sequential procedure relies only on the choice of $\alpha$. It does not introduce dependence on bandwidth into the SCAMP search procedure, implicitly or explicitly. The taut string, for example, requires setting the $\kappa$ parameter, or accepting its default value of $\kappa = 19$. 

In our experience, when applying SCAMP to real datasets the default $\kappa$-value of $19$ causes the taut string to produce very useful density estimates the vast majority of the time. However, in rare data sets we have found that the taut string can find a very large number of antimodes when applied to certain features. In these cases, exhaustive search of the annotation forest becomes impossible unless the feature is modified or removed from the analysis. 

To account for this rare circumstance, our implementation in the \pkg{scamp} package has a parameter to
limit the number of antimodes SCAMP will search along any coordinate in a given partition tree. In the 
event the taut string estimate exceeds the parameter value, the search terminates along that branch. In
our typical use cases, we set this parameter to $100$, though we expect this parameter should be
modified depending on the problem domain.

If we view an estimated density having a very large number of antimodes when in truth there are very few antimodes in the underlying distribution as an instance of type I error, we can think of switching to the sequential testing procedure as explicitly imposing a preference for type II error into the SCAMP search procedure in small sub-collections. The current implementation guarantees that SCAMP will never produce greater than $4$ modal sub-collections when examining small subsets of the data matrix, even in cases where there truly are more than $4$. Should more efficient implementations of the $N$-dip become available, this restriction can be reduced for larger number of modes in the $N$-dip and more than $400$ observations in a sub-matrix.

The recursive search terminates along a branch when it encounters a sub-matrix whose coordinates all have dip test $p$-values exceeding $\alpha$,
or the sub-matrix under consideration contains fewer than $m$ observations. 
In the former case, SCAMP has found a candidate cluster and the indices of the rows in the sub-matrix are recorded.
In the latter case, stopping is viewed as uninformative and the indices are deleted.

Ideally this search step is exhaustive. 
In practice, exhaustive search is impractical for data matrices with even a moderate number of
features.
Sampling is used to address this limitation: our implementation in the package \pkg{scamp} is able
to search uniformly at random among the space of partitional trees in the annotation forest until
finding a specified number of candidate clusters. We empirically justify this sampling approach with
our simulated data experiment in section \ref{section:casestudes}.

\subsubsection{Select Candidate Clusters}\label{scamp:phase3}

Clustering of the dataset begins with SCAMP imposing a \textit{preference values} on the set of 
candidate clusters collected on phase 2.
As discussed in section \ref{section:introduction}, the preference function reflects the users beliefs 
about components in the mixture \eqref{scamp:cluster_distribution}.
For its default, SCAMP prefers candidate clusters that are unimodal, symmetric, and have low variance 
along each coordinate projection.
If SCAMP is applied to a data matrix produced in a domain where other cluster shapes are desirable, the
preference function can be modified to select accordingly.

SCAMP uses 
the sample L-moments of \cite{hosking1990moments,hosking2007some} to assign preference
scores to candidate clusters.
Recalling definition \ref{section1:def:amcluster} from section \ref{section:introduction}, 
denote the set of $s$ candidate clusters found in SCAMP phase 2 by
\[
\mathcal{C} \equiv \left\{X_{\mathcal{I}\times p}^1,\ldots,X_{\mathcal{I}\times p}^r,\ldots,X_{\mathcal{I}\times p}^s \right\}\ ,
\]
with $1 \leq r \leq s$ and with the indices into the original data matrix different for each element of $\mathcal{C}$.
Denote Hosking's trimmed second, third, and fourth order
for the $i^{th}$  coordinate of $X_{\mathcal{I}\times p}^r$ by $\sigma_i^r, \tau_i^r$, and $\phi_i^r$  respectively, with $1 \leq i \leq p$. 
Additionally, denote the p-value of the dip statistic $\hat{D}_1$  for each of the $p$-coordinates of of $X_{\mathcal{I}\times p}^r$ by $\delta_i^r$.

For each fixed candidate cluster $X_{\mathcal{I}\times p}^r$ in $\mathcal{C}$, SCAMP assigns it a preference value by first computing
\begin{align}
&\delta^r \equiv \min_{1 \leq i \leq p} \delta_i^r \ ,\ \sigma_{\text{max}}^r \equiv \max_{1 \leq i \leq p} \abs{\sigma_i^r}  \ ,\ \sigma_{\text{sum}}^r \equiv \sum_{1 \leq i \leq p} \abs{\sigma_i^r} \ ,\nonumber \\
&\tau_{\text{max}}^r \equiv \max_{1 \leq i \leq p} \abs{\tau_i^r}  \ , \  \tau_{\text{sum}}^r \equiv \sum_{1 \leq i \leq p} \abs{\tau_i^r} \ ,\ \phi_{\text{max}}^r \equiv \max_{1 \leq i \leq p} \abs{\phi_i^r} \ ,\ 
\text{ and } \phi_{\text{sum}}^r \equiv \sum_{1 \leq i \leq p} \abs{\phi_i^r} \ .\nonumber 
\end{align}
With these quantities computed for each element of $\mathcal{C}$, SCAMP next computes
\begin{align}
&\sigma_{\text{max}} \equiv \max_{1 \leq r \leq s} \sigma_{\text{max}}^r  \ ,\ \sigma_{\text{sum}} \equiv \max_{1 \leq r \leq s} \sigma_{\text{sum}}^r \ ,
\tau_{\text{max}} \equiv \max_{1 \leq r \leq s} \tau_{\text{max}}^r  \ , \nonumber \\
&\ \tau_{\text{sum}} \equiv \max_{1 \leq r \leq s} \tau_{\text{sum}}^r \ ,
\phi_{\text{max}} \equiv \max_{1 \leq r \leq s} \phi_{\text{max}}^r  \ , \text{ and } \ \phi_{\text{sum}} \equiv \max_{1 \leq r \leq s} \phi_{\text{sum}}^r \ .\nonumber 
\end{align}
Using these maximums across $\mathcal{C}$, normalized values of the sample L-moments are then computed for each $X_{\mathcal{I}\times p}^r$: 
\begin{align}
\sigma_{nm}^r \equiv 1 - \frac{\sigma_{\text{max}}^r}{\sigma_{\text{max}}}  \ ,\ \sigma_{ns}^r \equiv 1 - \frac{\sigma_{\text{sum}}^r}{\sigma_{\text{sum}}} \ , \label{section:scamp:normstep}
\end{align}
along with the analogous quantities $\tau_{nm}^t, \tau_{ns}^r, \phi_{nm}^r,$ and  $\phi_{ns}^r$.
The preference score of cluster $X_{\mathcal{I}\times p}^r$ is then defined to be
\begin{align}
\mathcal{P}\left(X_{\mathcal{I}\times p}^r\right) \equiv  \delta_r + \frac{\sigma_{nm}^r+\sigma_{ns}^r}{2}+ \frac{\tau_{nm}^r+\tau_{ns}^r}{2}+ \frac{\phi_{nm}^r+\phi_{ns}^r}{2}\ .\label{section:scamp:preference}
\end{align}
The normalization step \eqref{section:scamp:normstep} allows the following interpretation of the preference score \eqref{section:scamp:preference}: the higher the better. 
Notice that $X_{\mathcal{I}\times p}^{r}$ enters $\mathcal{C}$  because $\delta_r > \alpha$. 
By using $\delta_r$ as a signal of unimodality in \eqref{section:scamp:preference}, we  SCAMP is encouraged to discriminate the degree of unimodality.
Suppose SCAMP is applied to a data matrix with $\alpha = 0.05$. 
Two hypothetical candidate clusters $X_{\mathcal{I}\times p}^{r_1}$ and $X_{\mathcal{I}\times p}^{r_2}$ enter $\mathcal{C}$ with $d_{r_1} = 0.06$ and $d_{r_2} = 0.94$. $\mathcal{P}$ 
on the basis of the dip alone, $\mathcal{P}$ is defined to prefer $X_{\mathcal{I}\times p}^{r_2}$.
The preference score \eqref{section:scamp:preference}  averages the trimmed sample L-moments to balance between the worst shape and average shape of the coordinates of a candidate cluster $X_{\mathcal{I}\times p}^{r}$. 
For example, $(\sigma_m^r+ \sigma_s^r)/2$ is meant to score a candidate cluster so that if either one single coordinate has extremely high spread or if many coordinates have moderate spread,
then SCAMP will not find the candidate cluster particularly appealing. Similar reasoning motivated $(\tau_m^r+\tau_s^r)/2$ and $(\phi_m^r+\phi_s^r)/2$.
The sample L-moments are trimmed in order to moderate the preference score penalty incurred by a candidate cluster when it contains an outlying observation along any of its coordinates.

By scoring each candidate cluster, SCAMP has created a set of weights
\[
\mathcal{P} 
\equiv \left\{
\mathcal{P}\left(X_{\mathcal{I}\times p}^1\right),
\ldots,
\mathcal{P}\left(X_{\mathcal{I}\times p}^r\right),
\mathcal{P}\left(\ldots,X_{\mathcal{I}\times p}^s\right) \right\}
\]
associated with the set of candidate clusters $\mathcal{C}$. A graph can now be induced by viewing each
element of $\mathcal{C}$ as a node, and inducing the adjacency matrix with edge set
\[
\mathcal{E} \equiv \left[ 
\mathbb{E}\left(X_{\mathcal{I}\times p}^{r_1},X_{\mathcal{I}\times p}^{r_2}\right)
\ \vline\ 
\mathbb{E}\left(X_{\mathcal{I}\times p}^{r_1},X_{\mathcal{I}\times p}^{r_2}\right) = 
\begin{cases}
0 \text{ if }  X_{\mathcal{I}\times p}^{r_1} \cap X_{\mathcal{I}\times p}^{r_2} = \varnothing  \\
1 \text{ otherwise} \\
\end{cases}
\right]\ .
\]
That is, two candidate clusters are connected by an edge if they share a common observation from the data matrix. 

Clustering of the data matrix proceeds by selecting a subset of $\mathcal{C}$ that maximizes the sum of the associated preferences $\mathcal{P}$  and  no two selected clusters are connected in $\mathcal{E}$.
This is an instance of maximum weight independent set (MWIS) integer program.
The MWIS problem is NP-hard; guarantees on the performance of approximate solutions are not known expect in special cases [\cite{brendel2010segmentation}].
SCAMP performs a greedy search: it sequentially picks the available candidate cluster with the highest preference score, 
and eliminates all candidate clusters connected in $\mathcal{E}$ from contention in subsequent selection steps. 

\subsubsection{Cluster residual observations}

%
%

The initial set of selected clusters, those candidate clusters picked by the greedy selection procedure of phase 3, often determine a partial clustering of the data matrix.
This is because the set of selected clusters consists of leaves of different partition trees in the annotation forest.
The residual observations that are not elements of any selected clusters do not necessarily obey the definition of an $\alpha$-$m$-cluster.
When these residual observations exhibit multimodality at level $\alpha$ along some of their coordinate projections, they are thought to represent samples from components of the mixture \eqref{scamp:cluster_distribution} that were not found in the initial search.
Therefore, a SCAMP iteration does not terminate after a single search-and-selection pass.

To complete the clustering of the data matrix, SCAMP conducts the search phase 2 and the selection phase 3 on the subset of rows of the data matrix corresponding to the residual observations.
This creates a new set of candidate clusters $\mathcal{C}$. The new set is scored according to the preference function $\mathcal{P}$. 
New clusters are then selected and appended to the clusters already selected in the earlier selection round.
Once again, the secondary selection procedure may produce residual observation.
So, the procedure recurses, applying the search and selection phases to smaller and smaller subsets of the original data matrix until either no residual observations appear, 
the residual observations comprise an $\alpha$-$m$-cluster, 
or there are fewer than $m$ residual observations.

At the completion of this process, a set of selected clusters $\mathcal{S}$ has been created. 
By its construction, $\mathcal{S}$ partitions the data matrix into $\alpha$-$m$-clusters. 
This is why SCAMP does not require the user to select the number of clusters as a tuning parameter: the number of clusters follows instead from the choice of $\alpha$ and $m$. 

$\alpha$ may initially seem as difficult a parameter to set as the number of clusters $k$.
In practice, however, we have found it relatively simple to set since $\alpha$ describes
a one-dimensional characteristic of the data matrix.
Our standard procedure is the following.
Before we apply SCAMP to a new data matrix for a given application, we first
decide on an absolute upper bound on $\alpha$. 
Intuitively, this corresponds to the minimum amount of evidence we require to split 
a coordinate into modal sub-collections.
Based on our experience of applying the dip test to a wide-range of datasets, we set this 
upper bound to a default value of $\alpha = 0.25$.

Next, we refine this upper bound on $\alpha$ by  
adding the structured noise of phase 1 to the columns of the data matrix, then computing the 
dip test p-values for each column vector, and finally plotting their distribution.
If there is a clear point of separation in the distribution of p-values that
is slightly larger than our default choice of $\alpha=0.25$, 
we can modify our choice upwards to respect the observed separation.
Our case studies in section \ref{section:casestudes} provide several examples 
of these plots and how we use them.

When there is no clear point of separation in the p-values but a subset of the p-values fall below
our default $\alpha$ value, we normally use the default.
However, if we want a sparser clustering of the data matrix, we 
next look at histograms of the features below our bound,
and pick an $\alpha < 0.25$ value large enough to ensure those features which appear 
multimodal from inspection are entered into the search phase.

In our experience, classical default values, such as $\alpha = 0.05$, can also work well for
high-dimensional datasets with a large number of observations, 
so long as the signal of interest is well-separated:
the separated mixture simulation of section \ref{section:casestudes} is one example of 
such a scenario. 
To summarize: while the choice of $\alpha$ can vary from problem to problem, 
inspecting the distribution of dip-test p-values for the available features provides a simple
data-driven approach for modifying this parameter from a default setting of $\alpha = 0.25$.

As for the parameter $m$, if prior knowledge informs the expected cluster size, we can use this
knowledge to set a value. If no prior knowledge informs the analysis, we simply pick as small a
value as computational resources allow: our default value is $m=25$.
Beyond the computational limit, there is an absolute lower bound of $m = 4$ due to SCAMP's
reliance on the dip test.

\subsubsection{Annotate selected clusters}

A SCAMP iteration concludes by assigning labels to the selected clusters $\mathcal{S}$. 
Since each selected cluster is an $\alpha$-$m$-cluster, we expect the distribution of observations within a cluster to be somewhat unimodal along any given coordinate.
Empirically we have observed that while the hypothesis of unimodality generally holds for clusters selected early in the selection process,
clusters selected near the end of the selection process can be asymmetric with shoulders.
Nevertheless, our annotation procedure behaves as if the assumption of unimodality holds equally well for all selected clusters.

To begin, annotation boundaries are determined for each of the coordinates of the data matrix $X_{k \times p}$ that are multimodal at level $\alpha$.
The annotation boundaries are computed using data collected in the annotation forest search of phase 2.  
For each of the $j$ coordinates of the data matrix, $1 \leq j \leq p$, SCAMP records the location of all cut-points found in all trees of the annotation forest grown during the \textit{initial}
candidate cluster search. It records each collection of cut-points in a separate set. Define
\begin{align}
  K_i^j \equiv \left\{(\kappa_1,\ldots,\kappa_i)  \ \vline \
  \begin{matrix}
    \text{SCAMP induces } i+1\text{ modal groups in coordinate }\ j\\
    \ \text{ with i cut points } \kappa_1 < \ldots < \kappa_i\\
  \end{matrix}
    \right\}\ , \label{scamp:cutPointDist}
\end{align}
and the cardinality of each set by $\mathcal{K}_i^j = \abs{K_i^j}$. Define $m^j \text{ to be the largest index } i \text{ such that } \mathcal{K}_i^j > 0$.  
The annotation set for the $j^{th}$ coordinate is $K_a^j$, with $a$ defined to be the smallest index $1 \leq a \leq m^j$ such that $\mathcal{K}_a^j  \geq \mathcal{K}_k^j$ for all $1 \leq k \leq m^j$.

Supposing the annotation set $K_a^j$ has cardinality $\mathcal{K}_a^j = n$,
the annotation boundaries $K^j$ are derived for the $j^{th}$ coordinate by computing the median cut-point 
for each of the $a$ annotation coordinates:
\begin{align}
K^j \equiv \left\{(k_1,\ldots,k_a)  \ \vline \ k_b \equiv \text{median}_{1 \leq i \leq n} \ \kappa_{(b,i)} \text{ for } 1 \leq b \leq a \right\}\ . \label{scamp:annotationBoundaries}
\end{align}
To annotate a specific coordinate in a selected cluster, three sample percentiles of the selected cluster are computed. 
By default the $q_l \equiv 50^{th} - \epsilon $, $q_m \equiv 50^{th}$, 
and $q_u \equiv 50^{th} + \epsilon$ percentile are used, 
leading us to annotate clusters relative to their median coordinate values.
These values can be adjusted by the user, depending how rigidly they want the final labels 
assigned to a cluster to respect the annotation boundaries along all labeled coordinates.
In our experience, this varies from problem to problem.
The label provided to each cluster is a proportion describing the relative position of the sample 
quantiles to the annotation boundaries \eqref{scamp:annotationBoundaries},
thereby associating the cluster with a mode of the coordinate projection.

In the special case $a=1$, the following occurs: if $q_l \geq k_1$, 
the cluster is given the label $1/1$; if $q_u \leq k_1$, the cluster is given the label $0/1$;
otherwise, the cluster is split, with points in the cluster below $k_1$ given the label $0/1$, 
and those above $k_1$ the label $1/1$.
If $a > 1$, then $q_m$ is compared to each of the boundaries $k_1,\ldots,k_a$.
If $q_m < k_1$, the cluster is labeled $0/a$; otherwise the cluster is labeled with the index of 
the largest annotation boundary $q_m$ exceeds.

To summarize: for each multimodal coordinate of the data matrix $X$, SCAMP clusters are assigned 
labels with fractions describing their relative position of the cluster column 
sample quantiles to annotation boundaries. 
In the \pkg{scamp} package, 
these fractions (up to $8$) are then mapped to a dictionary to give each 
observation an interpretable label for the particular fraction. 
For example, if $a=1$, SCAMP reports $0/1$ as ``lowest'', and $1/1$ as ``highest''; if $a=2$, 
$1/2$ is ``medium''; etc.  

\subsection{Multiple Iterations}

As noted at the start of this section, a single SCAMP iteration determines a partitional clustering of a data matrix $X_{k \times p}$.
The clustering produced by that iteration is always dependent on the realization of the random noise in section \ref{scamp:noiseStep}.
Further uncertainty is added if the search of section \ref{scamp:phase2} finds only a random subset of the annotation forest.

SCAMP is able to quantify the effect of these stochastic sections of the procedure by being run
multiple times on the same data matrix $X_{k \times p}$.
By providing a name for each selected cluster in a given iteration the annotation step, SCAMP also
provides names for each individual observation in the data matrix.
Because of this, we can track the naming history of each observation in the data matrix across
multiple iterations.
We emphasize the computation efficiency: 
keeping this record only requires updating a map for each observation across the iterations, 
rather than recording which observations are co-clustered as SCAMP iterates.
Labels assigned to a given observation define the keys for each map; the values are integer counts 
of the number of times the observations is labeled a particular way.

We view the name determined by a single SCAMP iteration  
as a vote for how a particular row of a data matrix should be described.
For some rows, the vote is consistent: every iteration SCAMP gives a specific
observation the same name. 
In such a case, we have high confidence the observation is named appropriately.
For other rows, the vote is split, and two or three names appear with some regularity. 
In these cases, it seems likely the appropriate name might be some combination 
of those names that appear regularly.
And for still other rows, the vote is inconclusive, with no collection of names appearing to be 
more than fluctuations of the SCAMP procedure itself.

Because of this, the output of the \textit{scamp} procedure in the package \pkg{scamp} is not one 
but \textit{two} clusterings of a data matrix $X_{k \times p}$.
The first is the clustering found by labeling each observation according to label it is given most
often across multiple observations.
The second is the clustering found by a run-off heuristic, which we now describe.

If, across multiple SCAMP iterations, an observation receives the same label in more than half of
them, the observation is named according to that label.
However, if an observation is given a label less than half but more than $30\%$ of the time, and
the second most frequent label appears more than $20\%$ of the time,
then the two labels are combined according to the following heuristic. If the most frequent label
describes a coordinate, but the second most frequent label does not, 
the final label is given the most frequent label's description. However, if the most frequent
label describes a coordinate by the ratio $a/b$  and the second most frequent label 
described it as $c/d$, that coordinate is then called $(ad + bc)/(2bd)$. For example, if the most
frequent label is $0/1$ (corresponding to the annotation ``lowest''), and the second most frequent
label is $1/2$ (corresponding to the annotation ``medium''), the combined label would become
$(0\cdot 2 + 1\cdot 1)/(2\cdot 1 \cdot 2) = 1/4$ (corresponding to the annotation ``medium-low'').

A similar combination heuristic is carried out in the case where the most frequent label for an observation occurs between $20\%$ to $30\%$ of the time, and the second and third 
most frequent more than $15\%$. 
Of course, other voting schemes could be considered in place of these heuristics, and they carry
interesting questions of their own: is any one voting scheme optimal, or does it depend on problem
domain? We note that instead of running SCAMP for a pre-specified number of iterations, SCAMP
could instead run until subsequent iterations produce 
no change (larger than $\epsilon$) in the clustering determined by the selected voting scheme. 
In our own use of SCAMP, we most-often use the result of the maximum-vote heuristic after setting
the number of iterations to as large a value as computation time allows.

\section{Simulations and Examples}\label{section:casestudes}

In this section we apply SCAMP to a variety of data sources, 
both real and simulated.
We use the adjusted rand index [\cite{rand1971objective,hubert1985comparing}] 
as one measure to quantify clustering performance, and refer to it by ARI.
We use the VI distance of \cite{meilua2007comparing} as another. 

Two independent clusterings of the same data matrix 
have an ARI value of $0$ in expectation. Two
identical clusterings have an ARI value of $1$.  
\cite{meilua2007comparing} notes that the
operative base-line of the ARI varies from $0.05$ to $0.95$ (pg. 876) and may be negative when
comparing certain clusterings (pg. 886). 
On the other hand, \cite{meilua2007comparing} proves that
the VI distance is a metric on the space of clusterings of a data matrix, ranges in value between
$0$ (when clusterings are identical) and $\log n$ (with $n$ the number of observations), and can
be normalized to compare clusterings across data matrices with different number of observations. 
We report both values when analyzing real data matrices. In our simulation studies, we report only
the ARI to save space in graphical summaries.

Unless otherwise noted, SCAMP uses its default values $\alpha=0.25$, $m=25$, and $\gamma=4$
in all case studies.

\subsection{Iris Data}

We begin by showing how SCAMP clusters the iris data set of \cite{fisher1936use}. 
The data matrix contains measurements of sepal length, sepal width,
petal length, and petal width for 150 irises of three types: setosa, versicolor, and virginica.
There are multiple ties due to rounding: only 43 petal length measurements are unique, 
the maximum of the four variables. 
Because of the numerous ties SCAMP's noise phase has a substantial effect on the clustering.

\begin{figure}[H]
  \centering
  \includegraphics[width=0.95\textwidth,keepaspectratio]{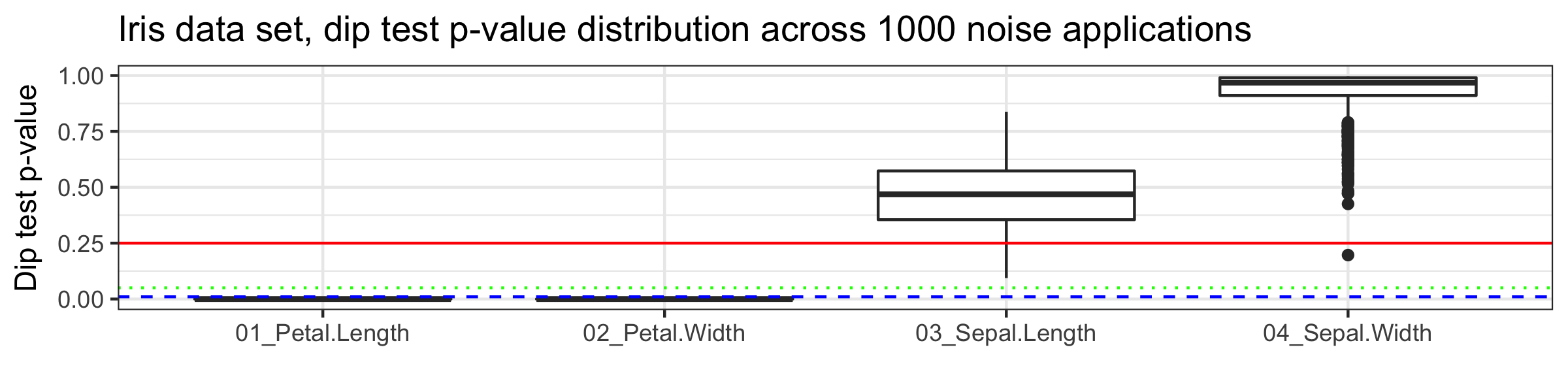}
  \caption[Pvalues]{The iris data set features, ranked according to their mean 
  dip test p-value across 1000 noise applications. We see that 
  the p-value of sepal length and sepal width are sensitive to SCAMP's noise phase.
      The plot shows classical threshold such as $\alpha =0.01$ (the dashed blue line), $\alpha = 0.05$
    (the dotted green line) will label the same two features as a more liberal threshold of $\alpha = 0.25$ (the solid red).}\label{figure:irisPvalue}
\end{figure}

Before running SCAMP, we first look at the distribution of dip test p-values for the data matrix.
Figure \ref{figure:irisPvalue} shows this distribution.
Based on this distribution, we see that choosing smaller values of $\alpha$ will not change
which columns are annotated: petal length and petal width have dip test p-values below $0.01$
every iteration.
We note, however, that more conservative $\alpha$ values will lead
to a sparser annotation forest, since the same $\alpha$ value is used to split a coordinate across
all depths of a partition tree.
We proceed with our default value of $\alpha = 0.25$. 

\begin{figure}[H]
  \centering
  \includegraphics[width=0.95\textwidth,keepaspectratio]{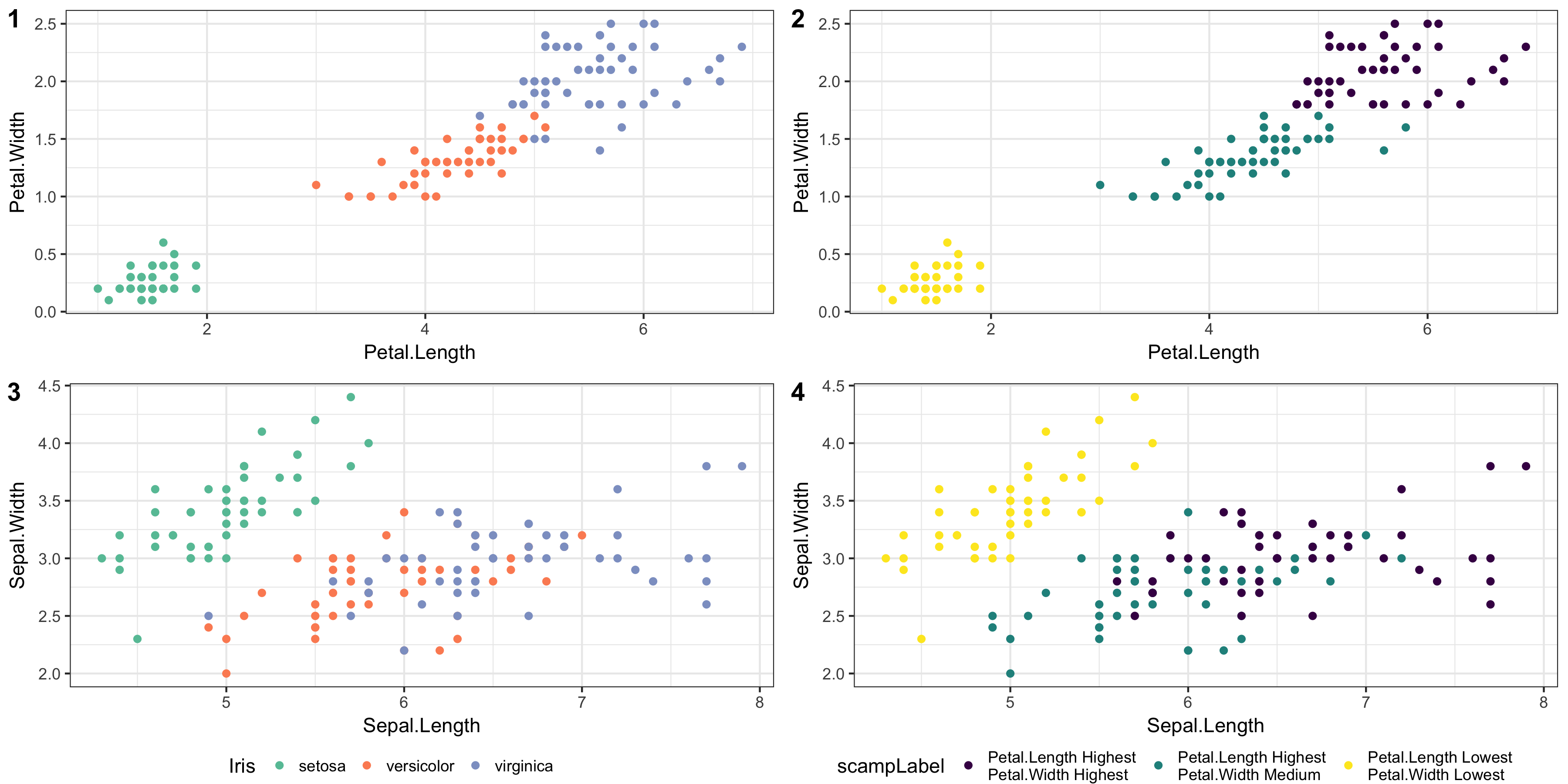}
  \caption[Iris data]{True labels and SCAMP labels of Fisher's iris data according to the maximum 
  vote across 500 iterations.
  The maximum vote SCAMP clustering ARI of $0.886$ 
  and an unadjusted VI distance [\cite{meilua2007comparing}] of $0.410$.
  SCAMP determines its labels automatically.}\label{figure:irisFigure}
\end{figure}

Figure \ref{figure:irisFigure} summarizes the clusterings found across 500 SCAMP iterations with
with $\alpha=0.25$, $m=25$, and the Gaussian noise parameter \eqref{scamp:GaussianNoise} 
$\gamma = 4$. 
Each iteration, SCAMP randomly samples $200$ candidate clusters from the annotation forest.
The cluster labels only describe an observations petal width and petal length, 
since sepal width and sepal length appear unimodal to SCAMP at level $\alpha=0.25$. The labels 
determined by SCAMP appear useful in characterizing the differences between the three iris 
species.

\subsection{Olive Oil}

We next apply SCAMP to the olive oil dataset of \cite{forina1983classification}.
This dataset has been previously analyzed by \cite{tantrum2003assessment}, \cite{stuetzle2003estimating}, and
\cite{chen2016comprehensive}.
We include this example to show how the descriptive labels produced by SCAMP can be useful for data
analysis, since the derived cluster labels explain differences between observations in different SCAMP clusters.

A description of these data is provided in the R package \pkg{classifly} [\cite{wickham2014classifly}],
from which we quote:
\begin{displayquote}
The olive oil data consists of the percentage composition of 8 fatty acids 
(palmitic, palmitoleic, stearic, oleic, linoleic, linolenic, arachidic, eicosenoic) 
found in the lipid fraction of 572 Italian olive oils. 
There are 9 collection areas, 4 from southern Italy (North and South Apulia, Calabria, Sicily), 
two from Sardinia (Inland and Coastal) and 3 from northern Italy (Umbria, East and West Liguria).
\end{displayquote}

The 9 collection areas constitute 3 production regions in this dataset: 
southern Italy, Sardinia, and northern Italy.

\begin{figure}[H]
  \centering
  \includegraphics[width=0.95\textwidth,keepaspectratio]{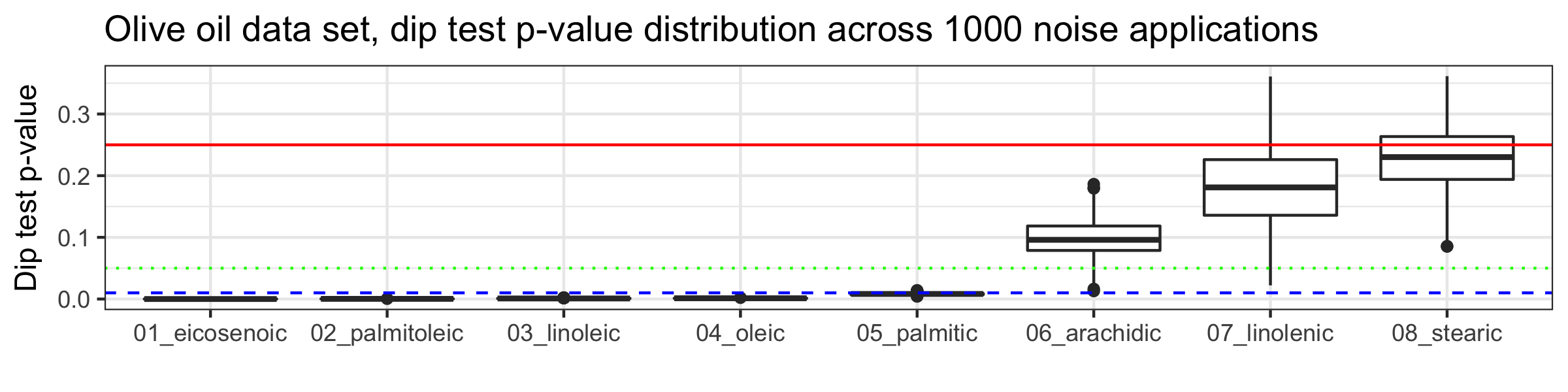}
  \caption[Pvalues]{The olive oil data set features, ranked according to their mean
  dip test p-value.
    The plot shows classical threshold such as $\alpha =0.01$ (the dashed blue line), $\alpha = 0.05$
    (the dotted green line) will label the same five features. Here, the default threshold of $\alpha = 0.25$ (the solid red)
    will inconsisently set labels for all eight features.}\label{figure:olivePvalue}
\end{figure}

Before running SCAMP, we again check our default choice of $\alpha$ against the
the data matrix: figure \ref{figure:olivePvalue} shows the distribution dip test 
p-values of the features in the data matrix.
We see the top 6 features always have a dip test p-value below our default $\alpha=0.25$
aross $1000$ noise iterations.
We observe, however, that the stearic fatty acid feature has a dip test p-value slightly above 
$0.25$ almost $35\%$ of the noise iterations. 
This poses a problem for SCAMP's annotation phase, since it indicates the stearic fatty acid feature
will often drop out of the cluster annotation strings. 
This in turn will bias the maximum vote annotation heuristic.

Because of this, we decide to increase our default $\alpha$: we 
run $5000$ SCAMP iterations with $\alpha=0.30$, $m=25$, and $\gamma=4$. 
Each iteration, SCAMP randomly samples $400$ candidate clusters from the annotation forest.
The maximum-label clustering heuristic determines $11$ clusters after $5000$ iterations.
Of these $11$ clusters, only $8$ have more than $25$ observations, our value for $m$.
We emphasize SCAMP can produce clusters with fewer than $m$ elements since the final 
clustering is determined by the most frequently assigned
label for each observation across the $5000$ SCAMP iterations.
The parameter $m$ only restricts cluster size in 
the search for candidate clusters for a single iteration. 

\begin{figure}[H]
  \centering
  \includegraphics[width=0.95\textwidth,height=0.5\textheight]{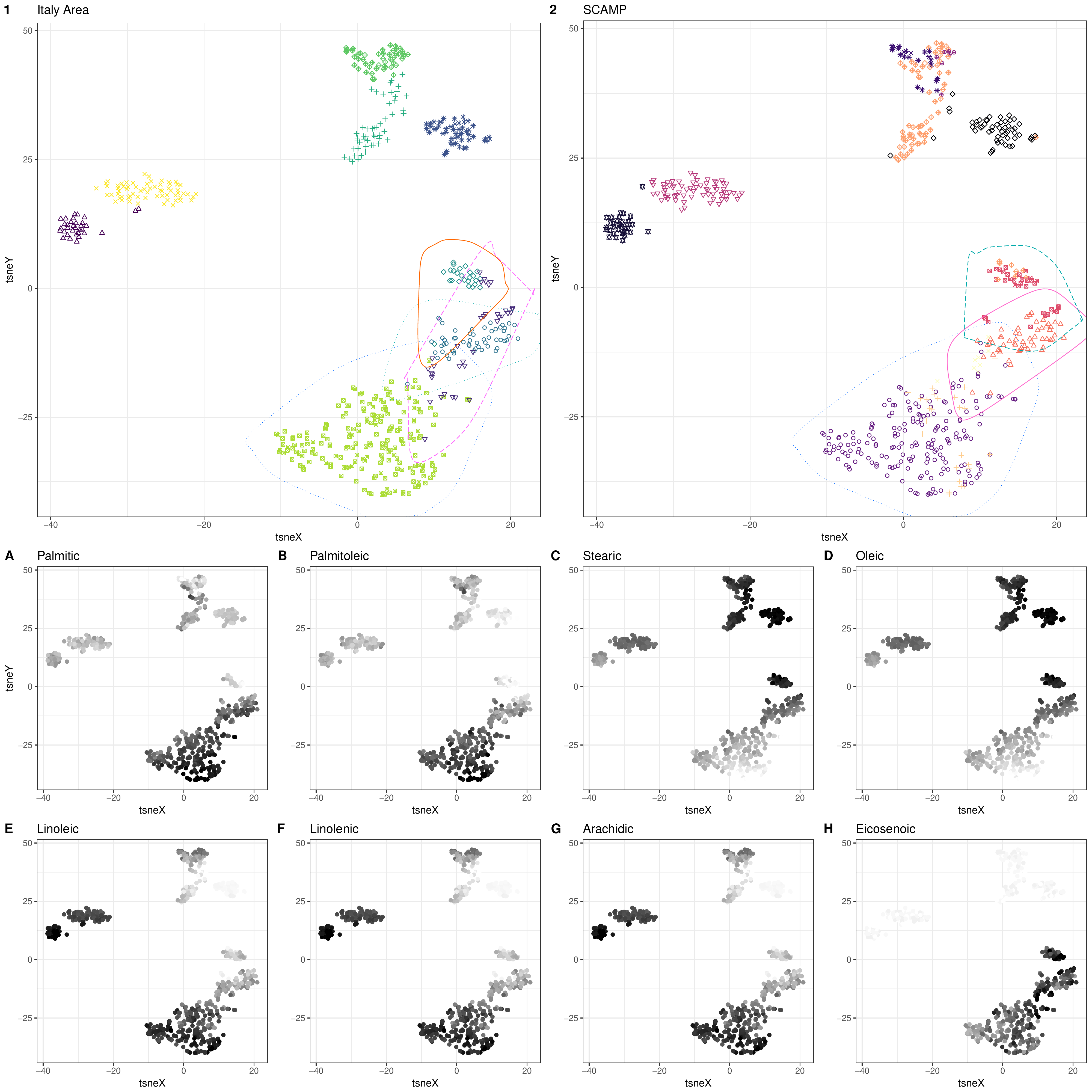}
  \caption[Olive oil visualization]{Each panel displays the same  t-SNE map of the olive oil dataset.
    Panel 1, titled ``Italy Area'', colors the points according to their region of production in Italy.
    Bounding boxes are drawn around four regions:
    the dotted-blue box bounds observations from South-Apulia;
    the solid-orange box bounds North-Apulia;
    the dashed-pink box bounds  Sicily;
    the final dotted-green box bounds Calabria.
    Panel 2, titled ``SCAMP'', colors the same t-SNE map according to an observation's membership in a
    SCAMP run-off cluster.
    Bounding boxes are drawn around three cluster:
    ``palmitic highest, palmitoleic highest, stearic lowest, oleic lowest, linoleic highest, linolenic  highest, arachidic highest, eicosenoic highest'' is bounded by
    the dashed-blue box;
    ``palmitic highest, palmitoleic highest, stearic medium-high, oleic lowest, linoleic lowest, linolenic highest, arachidic highest, eicosenoic highest'' is bounded by the solid-pink box;
    `palmitic lowest, palmitoleic lowest, stearic medium-high, oleic highest, linoleic lowest, linolenic highest, arachidic highest, eicosenoic highest'' is bounded by
    the long-dashed-green box.
    The remaining panels show the t-SNE map colored by the relative magnitude of the fatty-acid measurement:
    the darker the point, the higher the measurement.
    }\label{figure:oliveoil}
\end{figure}

Using the 9 Italian collection areas as the true cluster assignment,
the maximum-label
SCAMP clustering has an ARI of $0.734$ and unadjusted VI distance of $1.377$. 
Each cluster is labeled according to the relative quantities of the $8$ 
measured fatty acids.
For example, one cluster of $190$ observations is labeled 
``palmitic highest, palmitoleic highest, stearic lowest, oleic lowest, linoleic highest, linolenic 
highest, arachidic highest, eicosenoic highest''. A second cluster of $57$ observation is labeled
``palmitic highest, palmitoleic highest, stearic medium-high, oleic lowest, linoleic lowest, linolenic
highest, arachidic highest, eicosenoic highest''. A third cluster of $34$ observations is
labeled ``palmitic lowest, palmitoleic lowest, stearic medium-high, oleic highest, linoleic lowest, 
linolenic highest, arachidic highest, eicosenoic highest''.
These clusters broadly correspond to the southern Italy production region, which
contains the collection areas  South-Apulia (206 observations), North-Apulia (25 observations), Sicily (36 observations), and Calabria (56 observations).
We visualize these clusters in figure \ref{figure:oliveoil} using t-SNE [\cite{maaten2008visualizing}] to map the eight dimensions to two.

By coloring the t-SNE map according to the relative magnitude of each olive 
oil's fatty acid content, we see that the SCAMP labels reflect differences in the 
underlying fatty acid measurements. For example, olive oils in the cluster ``palmitic 
highest, palmitoleic highest, stearic lowest, oleic lowest, linoleic highest, linolenic
highest, arachidic highest, eicosenoic highest'' are mostly concentrated in a region of
the t-SNE map with high and low measured values of these fatty-acids, respectively (relative to the 
data set). A similar observation holds for the other clusters determined by SCAMP. 

From a data analysis standpoint, this allows an analyst to immediately understand why 
a given olive oil ends up in a particular SCAMP cluster. 
For example, within the South Italy production region, the three main 
SCAMP clusters all contain olive oils with
relatively high amounts of linolenic, arachidic, and eicosenoic fatty acids. 
The two SCAMP clusters that correspond (largely) to South-Apulia and Calabria also have
olive oils with 
high amounts of palmitic and palmitoleic acid, while the SCAMP cluster corresponding
(largely) to North-Apulia and Sicily has olive oils with
low amounts of these same fatty acids. On the other hand,
the olive oils in
the SCAMP clusters corresponding to
North-Apulia and Calabria have high amounts of stearic and 
linoleic acides, while the olive oils in the
South-Apulia/Sicily cluster has low amounts of these same fatty acids. 


\subsection{Separated mixtures}

Here we conduct a simulation study to model the following setting: an experimenter has
collected $n$ observations for $p$ variables. The experimenter wishes to cluster the data 
in order to find sub-collections of rows that exhibit a common signal along some sub-collection 
of the $p$ columns. 
To generate a matrix with such a property, we sample each sub-collection 
of rows from a multivariate distribution.
In the simplest case, the multivariate distribution is a Gaussian. 
Both the mean vector and covariance matrix differ from cluster to cluster.
In this simplest case, "no signal" corresponds to a mean vector entry $0$ and an
activation "signal" corresponds to a mean vector entry larger than $0$.

We examine three general scenarios.

\begin{enumerate}
\item{Scenario 1: Moderate sample size, moderate number of clusters.}
\item{Scenario 2: Large sample size, moderate number of clusters.}
\item{Scenario 3: Large sample size, large number of clusters.}
\end{enumerate}

\begin{figure}[H]
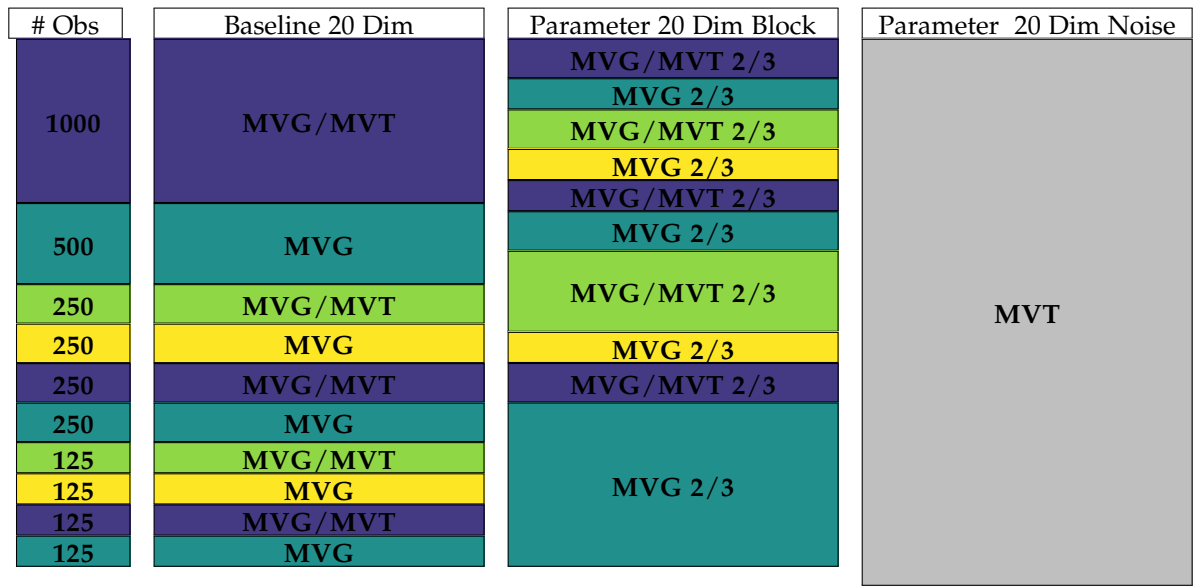

  \centering
  \[
    \raisebox{4\baselineskip+.5\myoffset}{
      \def\stackalignment{l}
      \stackunder[0pt]{\colblock[blue!0]{1}{3.5}{\text{\#\ Obs}}}
      {
        \stackunder[0pt]{\colblock[graphNode2]{5}{3.5}{\mathbf{1000}}}
        {\stackunder[0pt]{\colblock[graphNode4]{2.5}{3.5}{\mathbf{500}}}
          {\stackunder[0pt]{\colblock[graphNode6]{1.25}{3.5}{\mathbf{250}}}%
            {\stackunder[0pt]{\colblock[graphNode7]{1.25}{3.5}{\mathbf{250}}}
              {\stackunder[0pt]{\colblock[graphNode2]{1.25}{3.5}{\mathbf{250}}}
                {\stackunder[0pt]{\colblock[graphNode4]{1.25}{3.5}{\mathbf{250}}}
                  {\stackunder[0pt]{\colblock[graphNode6]{1}{3.5}{\mathbf{125}}}
                    {\stackunder[0pt]{\colblock[graphNode7]{1}{3.5}{\mathbf{125}}}
                      {\stackunder[0pt]{\colblock[graphNode2]{1}{3.5}{\mathbf{125}}}
                        {\colblock[graphNode4]{1}{3.5}{\mathbf{125}}}}}}}}}}}
      }
      {
        \stackunder[0pt]{\colblock[blue!0]{1}{10}{\text{Baseline\ 20\ Dim}}}
        {\stackunder[0pt]{\colblock[graphNode2]{5}{10}{\mathbf{MVG/MVT}}}
          {\stackunder[0pt]{\colblock[graphNode4]{2.5}{10}{\mathbf{MVG}}}
            {\stackunder[0pt]{\colblock[graphNode6]{1.25}{10}{\mathbf{MVG/MVT}}}%
              {\stackunder[0pt]{\colblock[graphNode7]{1.25}{10}{\mathbf{MVG}}}
                {\stackunder[0pt]{\colblock[graphNode2]{1.25}{10}{\mathbf{MVG/MVT}}}
                  {\stackunder[0pt]{\colblock[graphNode4]{1.25}{10}{\mathbf{MVG}}}
                    {\stackunder[0pt]{\colblock[graphNode6]{1}{10}{\mathbf{MVG/MVT}}}
                      {\stackunder[0pt]{\colblock[graphNode7]{1}{10}{\mathbf{MVG}}}
                        {\stackunder[0pt]{\colblock[graphNode2]{1}{10}{\mathbf{MVG/MVT}}}
                          {\colblock[graphNode4]{1}{10}{\mathbf{MVG}}}}}}}}}}}}
      }
      {
        \stackunder[0pt]{\colblock[blue!0]{1}{10}{\text{Parameter\ 20\ Dim\ Block}}}
        {\stackunder[0pt]{\colblock[graphNode2]{1.25}{10}{\mathbf{MVG/MVT\ 2/3}}}
          {\stackunder[0pt]{\colblock[graphNode4]{1}{10}{\mathbf{MVG\ 2/3}}}
            {\stackunder[0pt]{\colblock[graphNode6]{1.25}{10}{\mathbf{MVG/MVT\ 2/3}}}%
              {\stackunder[0pt]{\colblock[graphNode7]{1}{10}{\mathbf{MVG\ 2/3}}}
                {\stackunder[0pt]{\colblock[graphNode2]{1}{10}{\mathbf{MVG/MVT\ 2/3}}}
                  {\stackunder[0pt]{\colblock[graphNode4]{1.25}{10}{\mathbf{MVG\ 2/3}}}
                    {\stackunder[0pt]{\colblock[graphNode6]{2.5}{10}{\mathbf{MVG/MVT\ 2/3}}}
                      {\stackunder[0pt]{\colblock[graphNode7]{1}{10}{\mathbf{MVG\ 2/3}}}
                        {\stackunder[0pt]{\colblock[graphNode2]{1.25}{10}{\mathbf{MVG/MVT\ 2/3}}}
                          {\colblock[graphNode4]{5}{10}{\mathbf{MVG\ 2/3}}}}}}}}}}}}
      }
      {
        \stackunder[0pt]{\colblock[blue!0]{1}{10}{\text{Parameter \ 20\ Dim\ Noise}}}
        {\colblock[light-gray]{16.5}{10}{\mathbf{MVT}}}
      }
    }
  \]
  \caption[Simulation Overview]{This graphical overview of simulation settings shows how different parameters affect the block-structure of a sampled data matrix.
    The number of observations refers to the baseline Scenario 1 in which there are 10 clusters.
    Each cluster in the baseline setting is sampled from a 20 dimensional multivariate Gaussian with randomly chosen mean vector and covariance matrix.
    Entries of the mean vector are constrained to be $0$ or $6$: an entry of $6$ indicates the signal is present in that cluster.
    Variances are bounded above by $3$.
    Simulation parameters govern the following:
    the clusters can be alternatively sampled from multivariate Gaussian and multivariate T;
    clusters can be transformed coordinate-wise by one of four maps;
    a second 20 dimensional block with additional cluster structure can be sampled;
    the mean vector of the second block can contain $2$ 
    components in $\{0,6\}$ or $3$ components in $\{0,3,6\}$;
    a third 20 dimensional block of $3000$ observations can be sampled 
    from a multivariate T with 5 degrees of freedom to model the inclusion of noise variables.
    This creates $32$ distinct scenarios, 
    creating a data matrix with variable numbers of columns depicted here.
    Color coding indicates which transformation affects the block. 
    The label on each block indicates the generating distribution and possible number of components.
  }\label{figure:simulationSettings}
\end{figure}

There are five binary parameters that can be used to modify a scenario.
Together they create $32$ distinct simulation settings within a scenario.
For each setting, we run multiple iterations of the simulation.
We provide a graphical overview here to illustrate how the simulation parameters change the underlying 
data matrix in figure \ref{figure:simulationSettings}.
The simulation parameters, and the modifications they determine, are described in detail in appendix 
\ref{section:appendixSettings}

\begin{figure}[H]
  \centering
  \includegraphics[height=0.35\textheight,keepaspectratio]{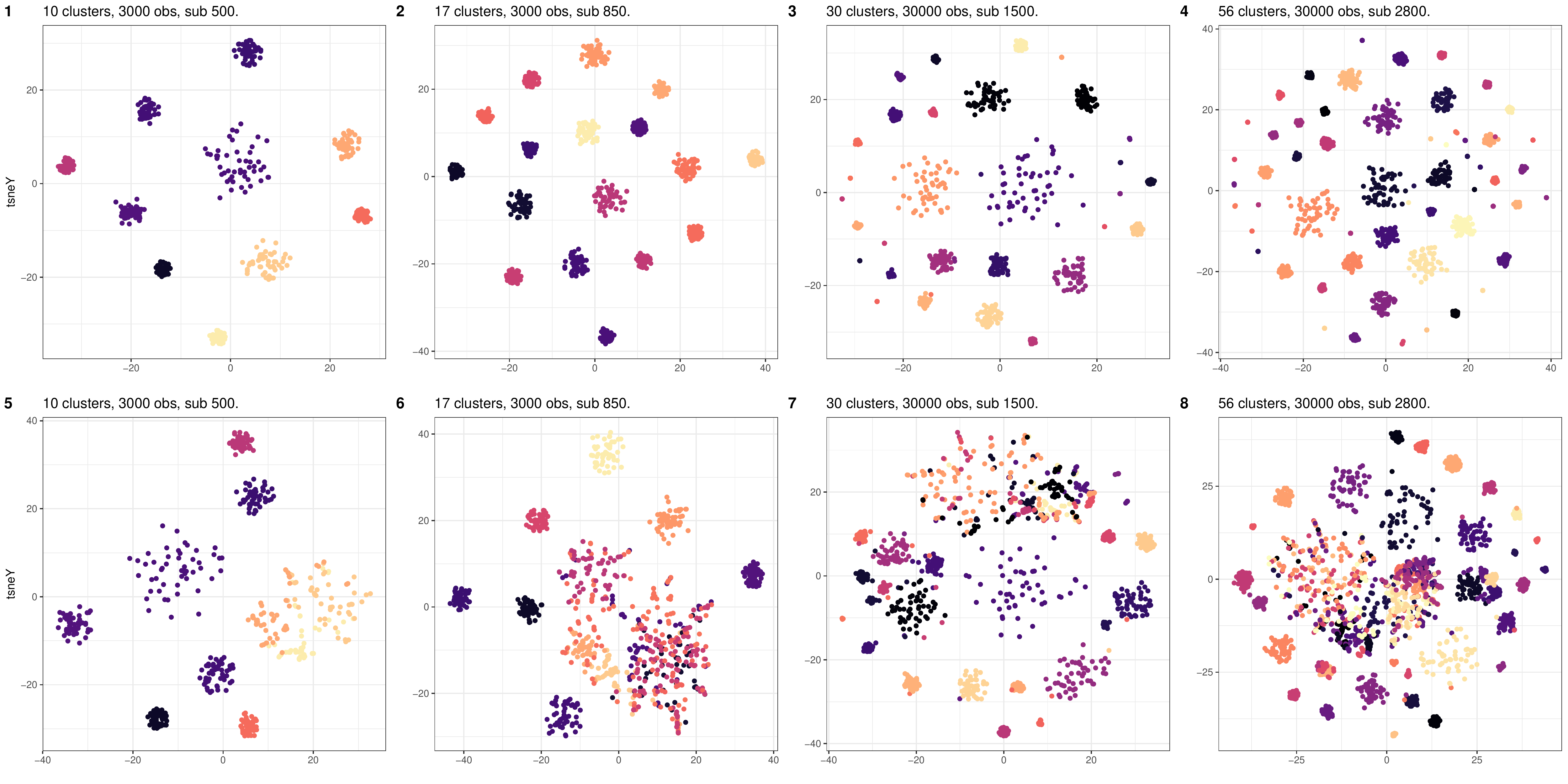}
  \caption[Simulation Matrix t-SNE]{t-SNE visualization of data matrices produced under eight different 
  simulation settings. 
    The number of observations in the data matrix, true number of clusters, and number of observations 
    sub-sampled for visualization are given in the label.
    Points are colored according to their cluster membership. 
    Each column in the display shows the cluster structure of a data matrix generated with identical 
    parameter settings, with one exception: 
    the diplays in the first row are untransformed, 
    while the clusters on the second row are taken through the maps described the appendix \ref{section:appendixSettings}}\label{figure:simulationTsneMatrixVix}
\end{figure}

In addition to the graphical depiction of the simulation parameters provided in figure \ref{figure:simulationSettings},
we also provide t-SNE visualizations of the cluster structure produced in
data matrices under different parameter settings in figure
\ref{figure:simulationTsneMatrixVix}.
t-SNE is particularly well suited to visualizing these data since many of the clusters 
are sampled from multivariate Gaussian or multivariate t distributions.
The t-SNE maps show significant overlap when the observations in clusters
are taken through the maps described in the appendix \ref{section:appendixSettings}.
In our experience, overlapping clusters are commonly observed in t-SNE maps produced 
from biological datasets.
Looking ahead, we see such overlap in our analysis of the GTEx data, visualized in figure \ref{figure:gtexTsne}.

\subsubsection{Scenario 1: Moderate sample size, moderate number of clusters}

The graphical overview of figure \ref{figure:simulationSettings} is specialized to this scenario.
In it, the data matrix contains $3000$ observations for all $32$ parameter settings.
There are either $10$ clusters in $20$ dimensions, or $17$ clusters in $40$ dimensions.
Cluster sizes range between $125$ and $1000$ observations in the $10$ cluster scenario.
The clustering ground truth is taken to be the label of the mixture component which produces an observation of the data matrix.

We sample data matrices $10$ times under each simulation setting. 
Each iteration, we apply the following clustering methods. 
Notice that several methods require the user to set the number of clusters as a parameter.
In the current simulation scenario, we provide all methods except SCAMP the number of clusters.

\begin{enumerate}
\item{(Oracle) Affinity Propagation (AP) [\cite{frey2007clustering,bodenhofer2011apcluster}]: number of clusters provided to the procedure ``apclusterK'' each iteration.
    Data matrix scaled to mean 0 and unit variance.
    We set the parameters ``maxits=2000'',``convits=2000'', and ``bimaxit=100''.}
\item{(Oracle) K-Means [\cite{hartigan1979algorithm}]: number of clusters set to the truth each iteration. Data matrix scaled to mean 0 and unit variance.}
\item{(Oracle) K-medoid [\cite{calinski1974dendrite,hennig2013find,pamkCite,clusterPAMCite}]: number of clusters set to the truth each iteration.
    Data matrix scaled to mean 0 and unit variance.}
\item{(Oracle) Model based clustering (Mclust) [\cite{fraley2002model,mclustcite2}]: number of components set to the truth each iteration.}
\item{SCAMP: we randomly search for candidate clusters, stopping after we find $50 \cdot \text{(number of columns)}$. We pre-set $\alpha=0.25$, $m=25$, and $\gamma=4$ across all simulations. We conduct a single SCAMP iteration per simulation setting and iteration. SCAMP
is provided $16$ threads to parallelize the search for candidate clusters.}
\item{SCAMP20: twenty iterations of the SCAMP procedure are performed and the maximum label heuristic used to cluster the data. Each iteration set to the same parameters as the single SCAMP run.}
\end{enumerate}


Observe that methods marked with the prefix ``(Oracle)'' have had the number of clusters  
provided to them. 
Looking ahead, the subsequent comparison of run-times omits the computation 
cost of estimating the number of clusters.
If the method uses the number of clusters as a tuning parameter, as with k-means,
we provide the number directly.
In the case of affinity propagation, 
parameters are modified to get the method close to the true number.
Model based clustering is capable of making its own estimate of the number 
of components present in a data matrix, 
but must be given a range of possible components to try.
A variety of methods could be used to estimate 
the number of clusters for k-means and k-medoids 
[see, for example, \cite{celeux1996entropy}, \cite{tibshirani2001estimating}, and 
\cite{dudoit2002prediction}].

\begin{figure}[!tbh]
  \centering
  \includegraphics[width=0.95\textwidth,keepaspectratio]{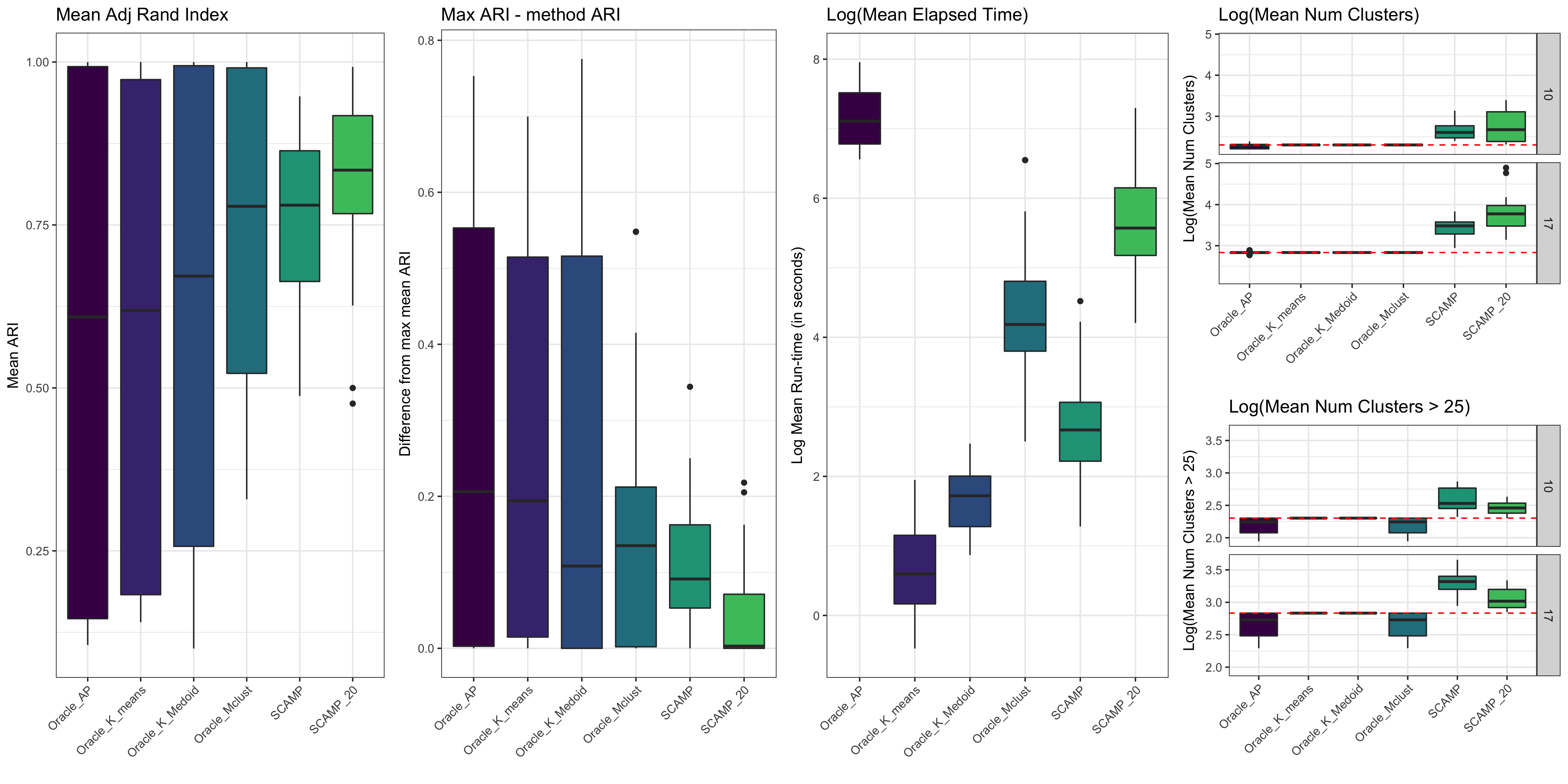}
  \caption[Simulation Study 1]{32 different simulations are run according to the parameters in table \ref{simulation:mixtureSettings}
    in appendix section \ref{section:appendixSettings}.
    50 iterations per setting. Mean of displayed statistics computed across iterations.
    Dataset has 3000 rows, and 20/40/60 columns depending on setting.
    True number of clusters either 10 or 17, depending on setting.
    Oracle prefix in method description indicates method given number of clusters or method parameters are adjusted to obtain that number.
    The logarithm of the true number of clusters is indicated by the dashed red line in each facet of the final panel.
    \label{figure:simulation01Figure}}
\end{figure}

Results from the simulation study are summarized in figure \ref{figure:simulation01Figure}.
The first panel displays the distribution of mean ARI over the $50$ iterations 
for each of the $32$ simulation settings.
There we can see that the methods with an oracle providing the number of clusters 
perform quite well when the simulation parameters 
match their implicit distribution assumptions:
K-means, K-medoids, and Mclust are able to recover cluster assignments almost perfectly in several 
of the multivariate normal settings.

However, there is serious downside risk evident for these same methods: when the parameters of the 
simulation introduce nuisance features and transform observations (through the maps 
\eqref{scamp:sim_settings}, described in the appendix),
the cluster assignments can differ substantially from the underlying mixture despite being provided
the true number of mixture components.
On the other hand, we see that SCAMP clusterings usually perform well in terms of ARI,
no matter the simulation settings: it is robust to distributional assumptions.
We also note that the twenty iteration of SCAMP improves the mean ARI under all parameter settings.

For each simulation setting, the method with maximum mean ARI is taken as a baseline value.
The difference between this best baseline ARI and each method's ARI is recorded.
The second panel of figure \ref{figure:simulation01Figure} shows the distribution of these 
differences across the $32$ simulation settings, with a zero value indicating the method performed 
best (according to adjusted rand index) in that setting.
This panel shows that a single SCAMP iteration has ARI within $0.1$ of the best performing method 
almost $50\%$ of the time. 
The panel also shows that the twenty SCAMP iterations are almost always performing as well as the 
best performing method, 
and performs the best in terms of adjusted rand index about $25\%$ of the time.

The third panel of figure \ref{figure:simulation01Figure} shows the distribution of the logarithm 
of mean run-time across methods.
We see that on these simulated data, SCAMP is relatively affordable when given $16$ threads: 
a SCAMP single iteration, which estimates the number of clusters, 
is usually finished after $20$ seconds.
The run-time comparison may be biased due to implementations: after reviewing the documentation
for the compared methods, we did not see the ability to provide them with additional execution 
threads.
We also see that multiple SCAMP iterations incur additional cost in terms of run-time.

The fourth panel shows the logarithm of the number of clusters estimated by each method 
in the top inset (the methods with an oracle are flat because they are given the truth).
The bottom inset shows the logarithm of the number of clusters with more than $25$ observations 
in them.
SCAMP estimate is usually overestimates the truth in both the $10$ and $17$ cluster scenarios.
Iterating twenty times improves the estimate at the cost of producing an increased number
of smaller clusters.

\subsubsection{Scenario 2: Large sample size, moderate number of clusters}

Our second simulation scenario explores settings with larger data matrices: our data matrix increases to $30000$ observations.
The graphical overview of figure \ref{figure:simulationSettings} 
can be modified to describe this setting by multiplying each cluster size by $10$.
We continue to use the 32 settings defined by the parameters described there.


We run a modified collection of methods in this setting due to the computational expense. They are:
\begin{enumerate}
\item{Leveraged Affinity Propagation (AP)[\cite{frey2007clustering,bodenhofer2011apcluster}]:
    Using the ``apclusterL'' implementation in the R package \pkg{apcluster}, 
    we set ``maxits=2000'',``convits=2000'', ``frac=0.1'' (corresponding to $10\%$ of the data), ``sweeps=5'' and ``q=0''.}
\item{(Oracle) Clustering Large Aplications [\cite{rousseeuw1990finding}]: The R implementation \textit{clara} in the package \pkg{cluster} is used, with $k$ set to the truth, the ``samples'' parameter set to $50$, and sampsize set to $3000$, which is $10\%$ of the rows of the data set.}
\item{(Oracle) K-Means: no change.} 
\item{(Oracle) K-medoid: no change.} 
\item{(Oracle) Model based clustering (Mclust):  we still provide the function ``Mclust'' with the number of clusters. We now initialize EM with $10\%$ of the data randomly selected.}
\item{SCAMP: no change.}
\item{SCAMP20: no change.}
\end{enumerate}

\begin{figure}[!tbh]
  \centering
  \includegraphics[width=0.95\textwidth,keepaspectratio]{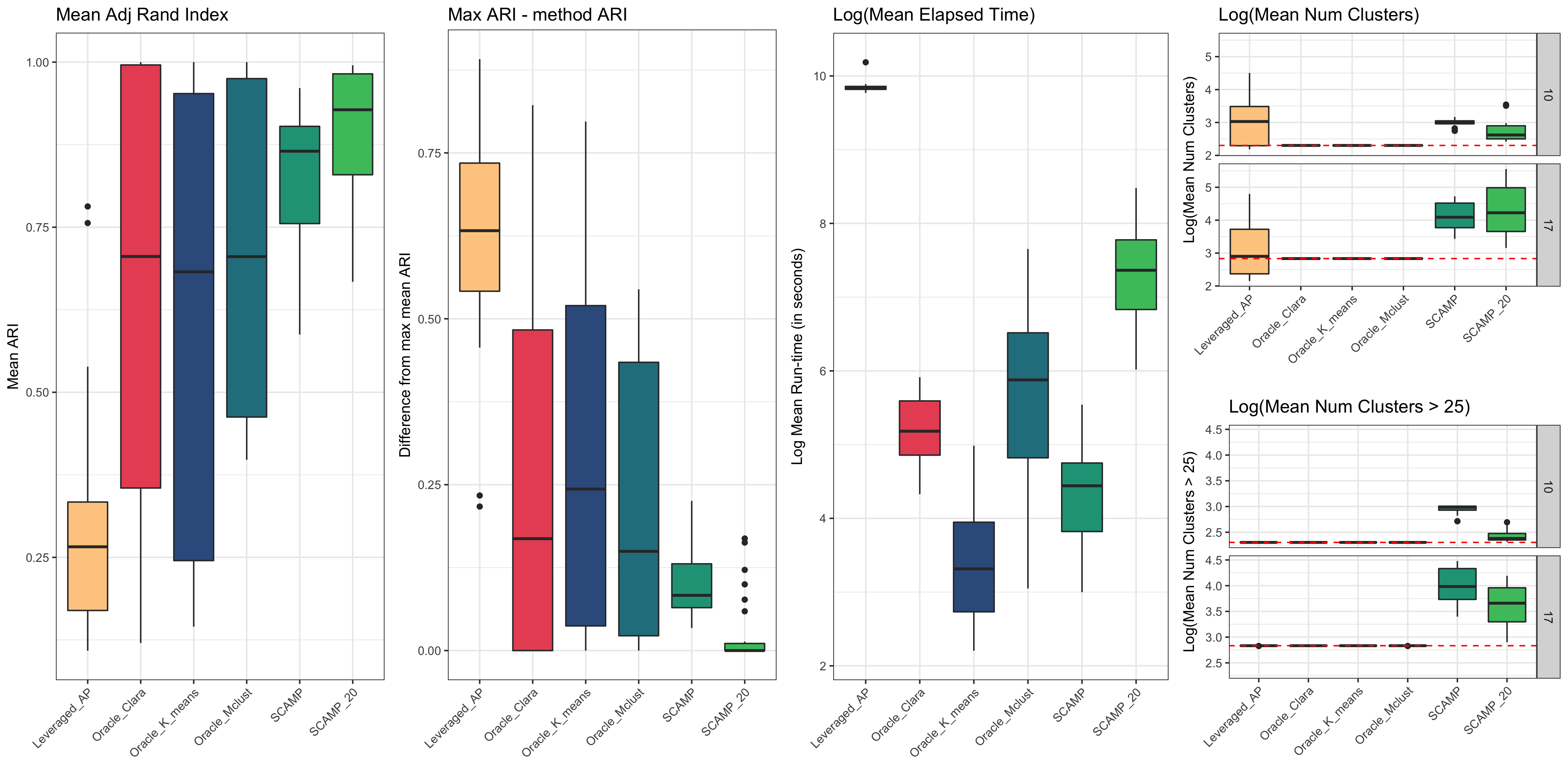}
  \caption[Simulation Study 2]{32 different simulations are run according to the parameters in table \ref{simulation:mixtureSettings}
    in appendix section \ref{section:appendixSettings}.
    10 iterations per setting. Mean of displayed statistics computed across iterations.
    Dataset has 30000 rows, and 20/40/60 columns depending on setting.
    True number of clusters either 10 or 17, depending on setting.
    Oracle prefix in method description indicates method given number of clusters or method parameters are adjusted to obtain that number.
    The logarithm of the true number of clusters is indicated by the dashed red line in each facet of the final panel.
    \label{figure:simulation02Figure}}
\end{figure}

For each of the $32$ simulation settings, we again run $10$ iterations per simulation setting.
The results of the simulation are displayed in figure \ref{figure:simulation02Figure}.
We interpret each panel of the figure as we did in figure \ref{figure:simulation01Figure}.

In the large matrix setting, we see that SCAMP performs well. 
Iterating SCAMP twenty times continues to improve performance at the cost of additional run-time.
On the other hand, the downside risk of alternative methods is more pronounced in this setting.
For some methods, this deterioration in performance in in part explained by their reliance on 
sub-sampling. Additionally, modifying the parameter settings for Leveraged Affinity Propagation
could potentially improve perfomance in terms of ARI -- we kept the current settings
for comparability to scenario 1.

\subsubsection{Scenario 3: Large sample size, large number of clusters}

We conclude by simulating data data with a large number of 
observations and a large number of clusters of different sizes.
We continue to generate a data matrix with $30000$ rows.
In this scenario, the basic mixture has 
30 components in $20$ dimensions, and $56$ components in the larger $40$ dimensions.
In the $30$ cluster composition, cluster sizes range between $100$ and $10000$ observations.
In the $56$ cluster composition, cluster sizes range between $50$ and $3950$.
Once again, for each of the $32$ simulation settings, 
we run $10$ iterations.
We refer the reader to the appendix for specific details.

In this scenario, we see that SCAMP continues to perform well. 
Since it does not have to sub-sample,  
it is able to recover much of component structure of the underlying mixture. 
In this scenario, the alternative methods almost all perform worse in terms of ARI than SCAMP.
SCAMP run for twenty iterations almost always performs best in terms of its adjusted rand index. 
A single SCAMP iteration compares well to other methods in terms of run-time,
and slightly over-estimates the true number of clusters on average.

\begin{figure}[!tbh]
  \centering
  \includegraphics[width=0.95\textwidth,keepaspectratio]{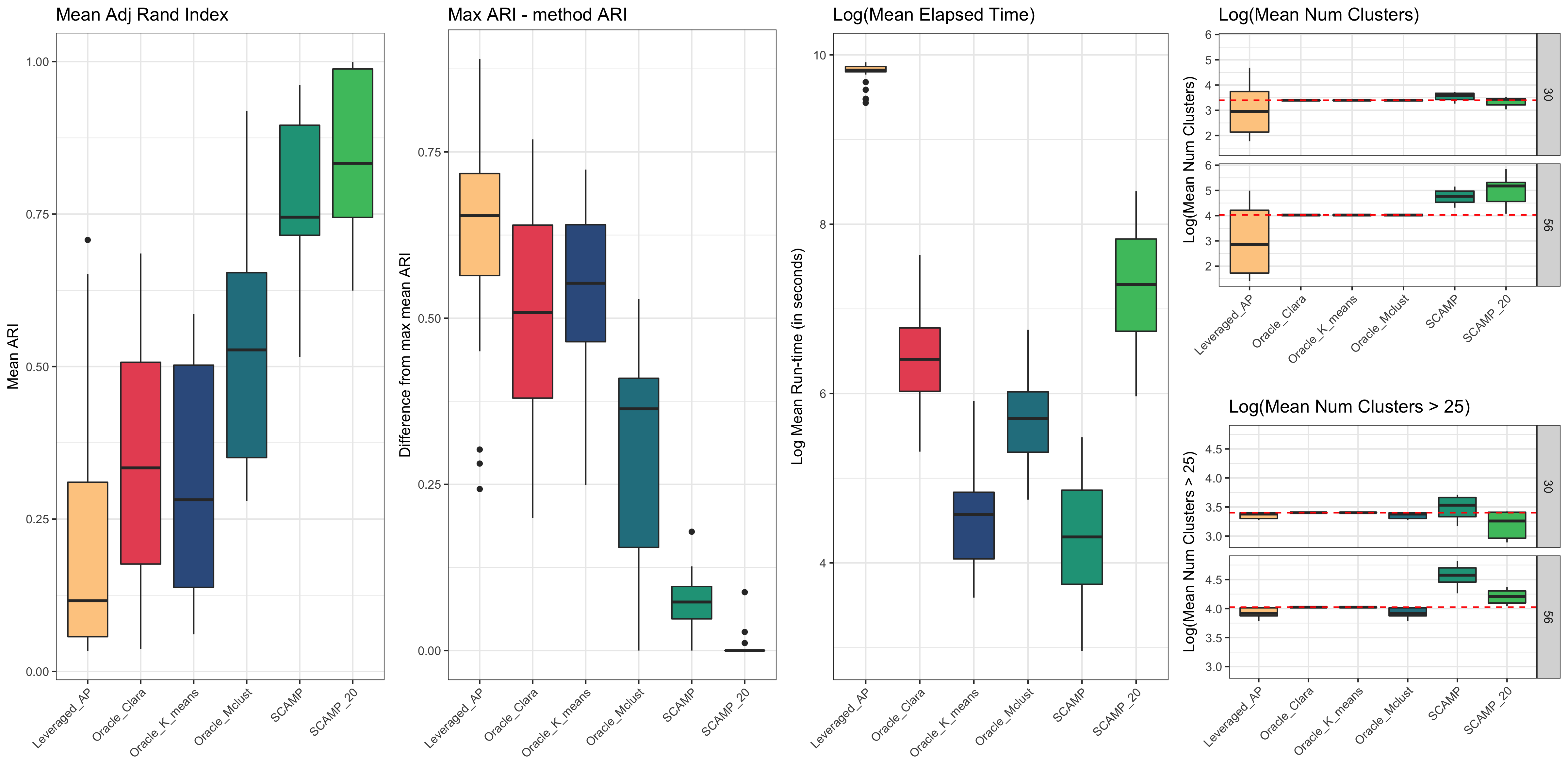}
  \caption[Simulation Study 3]{32 different simulations are run according to the parameters in table \ref{simulation:mixtureSettings} 
    in appendix section \ref{section:appendixSettings}.
    10 iterations per setting. Mean of displayed statistics computed across iterations.
    Dataset has 30000 rows, and 20/40/60 columns depending on setting.
    True number of clusters either 30 or 56, depending on setting.
    Oracle prefix in method description indicates method given number of clusters or method parameters are adjusted to obtain that number.
    The logarithm of the true number of clusters is indicated by the dashed red line in each facet of the final panel.
    \label{figure:simulation03Figure}}
\end{figure}

\subsection{Circular Data and Rotated Data}

The previous simulation of well-separated mixtures along the measurement axes is the ideal use case for SCAMP: in such cases, it is possible for SCAMP to provide interpretable clusterings of the data matrix. 
Here, we simulate two scenarios where SCAMP will perform poorly. 

\begin{figure}[H]
  \centering
  \includegraphics[width=0.95\textwidth,keepaspectratio]{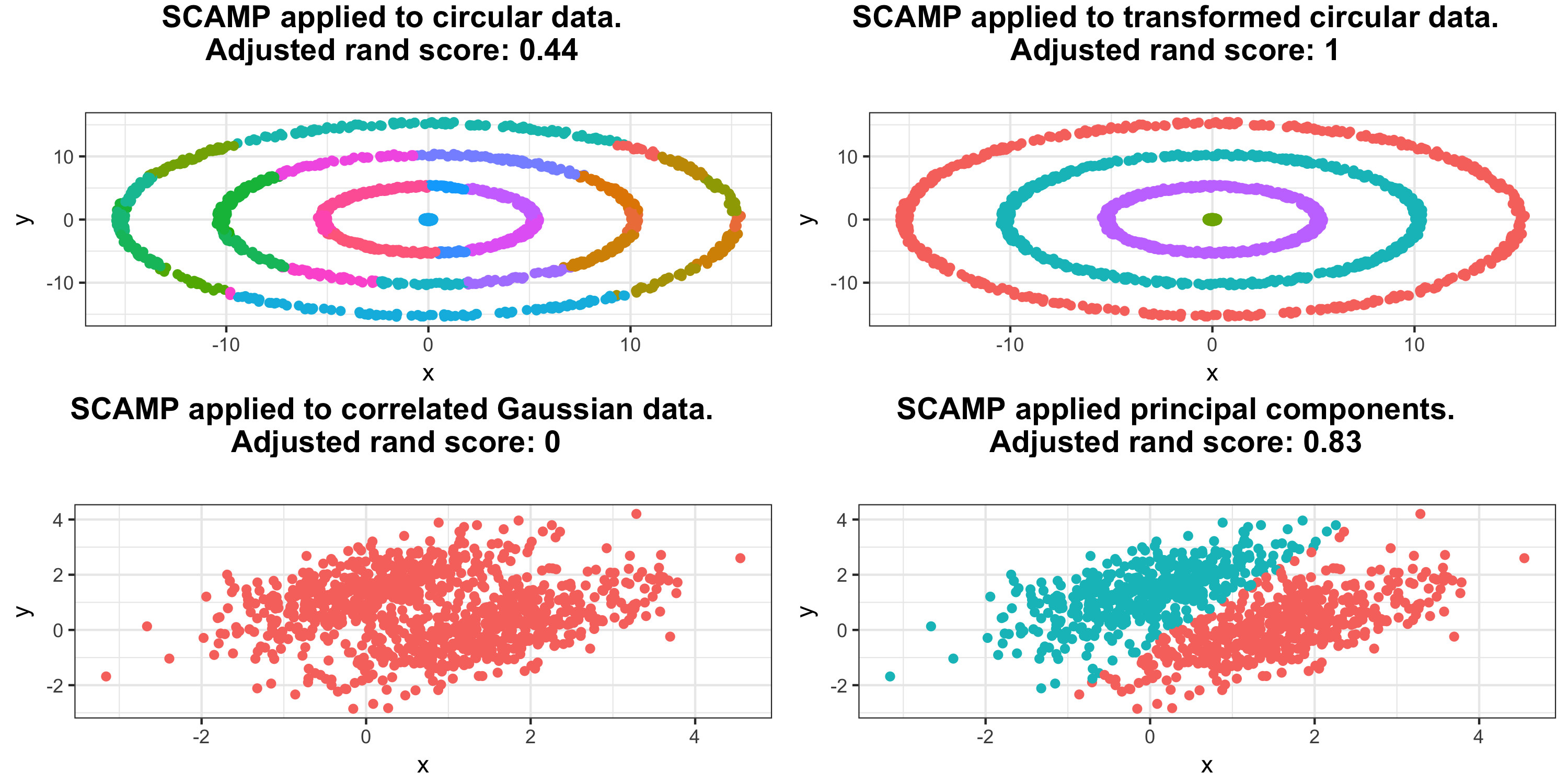}
  \caption[SCAMP Failures]{SCAMP clusterings of two simulated datasets. 
  In the case of circular data, SCAMP over partitions the data because the coordinate projections are highly multimodal. 
  After transforming the raw data through the map $\phi(x,y) \equiv x^2 + y^2$, SCAMP is able to capture the circular structure.
  In the case of data with cluster structure that is not visible along coordinate projects, 
  SCAMP under partitions and lumps all observations into a single cluster.
  After rotation through the principal components, SCAMP is able to recover group structure.}\label{figure:scampJoint}
\end{figure}

Figure \ref{figure:scampJoint} illustrates two examples in $\mathbb{R}^2$. 
In the first case, the data have circular structure. 
SCAMP over partitions these data since the coordinate projections appear highly multimodal, and so the connected structure of the data is lost. 
In the second case, the data are simulated from a mixture of two multivariate Gaussians with mean vectors $\mu_1 = (1.5,0)$, $\mu_2 = (0,1.5)$ and common covariance matrix 
$\Sigma \equiv
\begin{pmatrix}
1.0\ \ \ 0.7 \\
0.7\ \ \ 1.0 \\
\end{pmatrix}
$.
Such data have coordinate projections which appear unimodal. 
As a result, SCAMP clusters the data matrix by producing a single cluster consisting of the entire data matrix. 
The adjusted rand scores reflect SCAMP's difficulty with such data.

If one is willing to sacrifice interpretable clusterings, SCAMP can still be used to cluster 
these kinds of data matrices.
In the first case, SCAMP can be applied to the data matrix taken through 
the map $\phi(x,y) \mapsto x^2 + y^2$. 
In the second case, SCAMP can be applied to the principal components. 
Figure \ref{figure:scampJoint} shows how SCAMP clusters the data under these transformations. 
Since the clusterings are determined on transformed data matrices, the labels produced by SCAMP 
are not immediately interpretable on the original measurement scale.
However, these clusterings are still useful: SCAMP recovers the underlying number of circular 
components in the first case, the number of mixture components in the second case.
The adjusted rand scores are consequently improved.

In many clustering tasks, a method, such as k-means, is used as a clustering tool after the data 
have been transformed: PCA and kernel methods are two such examples.
These simulations show that in such work flows SCAMP might be an viable alternative. We conclude 
with a practical demonstration of such a work-flow.

\subsection{GTEx Data}

As a practical demonstration, we apply SCAMP to bulk RNA seq gene read count data from the 
the Genotype-Tissue Expression (GTEx) Project.
This project is supported by the Common Fund of the Office of the Director of the National Institutes of Health,
and by NCI, NHGRI, NHLBI, NIDA, NIMH, and NINDS.
The data used for the demonstration in this paper were obtained from the
\textcolor{blue}{\href{https://www.gtexportal.org/}{https://www.gtexportal.org/}}
from the v6p release in the the file ``GTEx\_Analysis\_v6p\_RNA-seq\_RNA-SeQCv1.1.8\_gene\_reads.gct.gz''
on April 19, 2018. 
An alternative analysis of the v6 data is given in \cite{dey2017visualizing}.

This dataset contains per-gene read counts for $56238$ genes across $8555$ samples.
The samples were collected from $450$ human donors.
Each of the $8555$ samples is taken from one of $31$ primary tissue types from a given donor.
The primary tissue type labels are further refined to $53$ tissue labels 
that describe the sample location.
For example $1259$ samples are given the primary tissue label ``Brain''. 
These $1259$ samples have $13$ corresponding
tissue labels ranging from ``Brain - Amygdala'' to  ``Brain - Substantia nigra''.
Here, will apply several clustering methods to the unlabeled count data (after transformation).
We will use these two label sets as versions of ground truth for clustering.

These count data are zero inflated and, for a given gene, 
differ by orders of magnitude across the samples.
To apply SCAMP to these data, we first remove $1558$ genes 
which have zero counts across all $8555$ samples.
This produces a data matrix of integer counts with $8555$ 
rows and $54680$ columns. We transform the counts for
each remaining gene by the map $\log_2(1+x)$.
We then normalize each gene across samples by the total sum of the transformed counts.

\begin{figure}[H]
  \centering
  \includegraphics[width=0.95\textwidth,keepaspectratio]{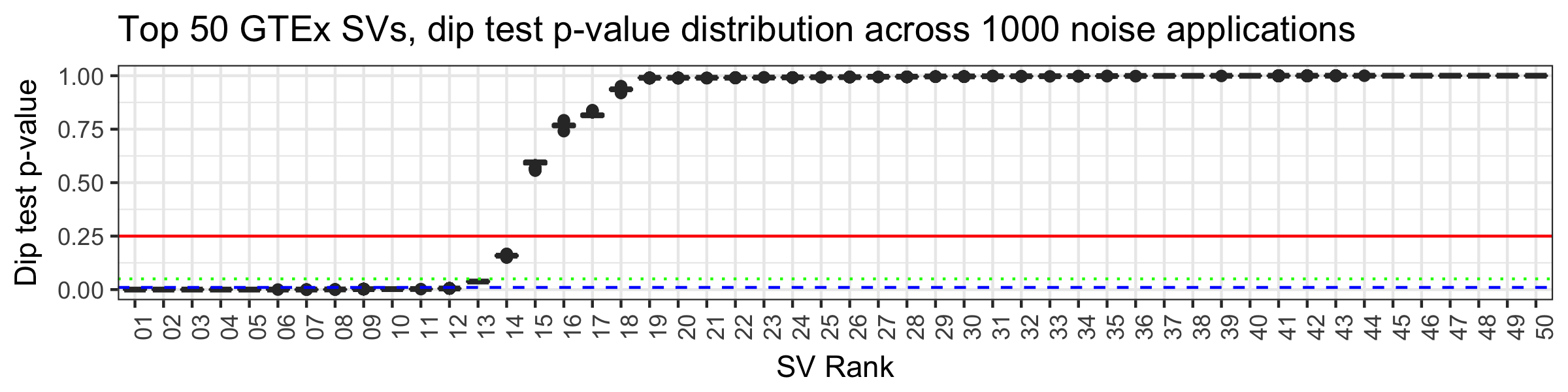}
  \caption[Pvalues]{The top 50 singular values for the pre-processed GTEx data, ranked according to their mean dip test p-value.
    The plot shows how the three thresholds for $\alpha$ select different numbers of SVs for
    annotation: $0.01$ (the dashed blue line) selects $12$ column vectors; $0.05$
    (the dotted green line) selects $13$; $0.25$ (the solid red) selects $14$.}\label{figure:gtexPvalue}
\end{figure}

To apply SCAMP, we next compute the top $50$ right-singular 
vectors of the data matrix and then use them rotate the transformed data matrix. 
We then apply SCAMP to the rotated data matrix.
Once again, 
the selection of $\alpha$ can be guided empirically before running SCAMP: 
figure \ref{figure:gtexPvalue} suggests our default value of $\alpha = 0.25$ 
will label clusters with all SVs exhibiting multimodality (according to the dip test) 
in these data.

For the purpose of demonstration, we will cluster the data with SCAMP 
using all three choices of $\alpha$ since the $\alpha$ values $\{0.01,0.05,0.25\}$
correspond to labeling clusters with $\{12, 13, 14\}$ SVs. 
For each choice of $\alpha$, we run SCAMP for 100 iterations.
Each search phase of each iteration,
we randomly sample $5000$ candidate clusters from the annotation forest.
The final clustering if found using the maximum-vote heuristic across the $100$ iterations.

We also cluster the dataset using affinity propagation, Mclust, k-means, and k-medoids.
We modify the parameters for affinity propagation to try to achieve convergence, setting 
``maxits'' to $5000$, ``convits'' to $5000$, and ``lam'' to $0.95$.
For Mclust, we let the number of mixture components 
range from $2$ to $100$, and select the final mixture using BIC: in this dataset, 
this selects 24 mixture components.
We use the estimate $24$ as the k parameter in k-means and k-medoids.
    
\begin{figure}[H]
\centering
\begin{tabular}{rlllllll}
  \hline
 & SCAMP 01 & SCAMP 05 & SCAMP 25 & AP & Mclust & k-means & k-medoid \\ 
  \hline
  Primary Tissue: VI & 1.296 & 1.362 & 1.453 & 1.383 & 1.003 & 1.205 & 1.017 \\ 
  Primary Tissue: ARI & 0.633 & 0.632 & 0.592 & 0.559 & 0.705 & 0.643 & 0.749 \\ 
  \# Primary Tissue & 31 & 31 & 31 & 31 & 31 & 31 & 31 \\ 
  Tissue: VI & 1.415 & 1.473 & 1.542 & 1.277 & 1.344 & 1.582 & 1.464 \\ 
  Tissue: ARI & 0.649 & 0.657 & 0.654 & 0.69 & 0.606 & 0.557 & 0.564 \\ 
  \# Tissue & 53 & 53 & 53 & 53 & 53 & 53 & 53 \\ 
  Method \# of Clusters & 54 & 61 & 75 & 41 & 24 & 24 & 24 \\ 
   \hline
\end{tabular}
    \caption[Score]{The VI distance and ARI for methods using the primary tissue label 
    and tissue label as ground truth.
    The number trailing each SCAMP column indicates the value of $\alpha$ used to cluster the data. }\label{figure:gtexScores}
\end{figure}

\begin{figure}[H]
  \centering
  \includegraphics[width=0.95\textwidth,keepaspectratio]{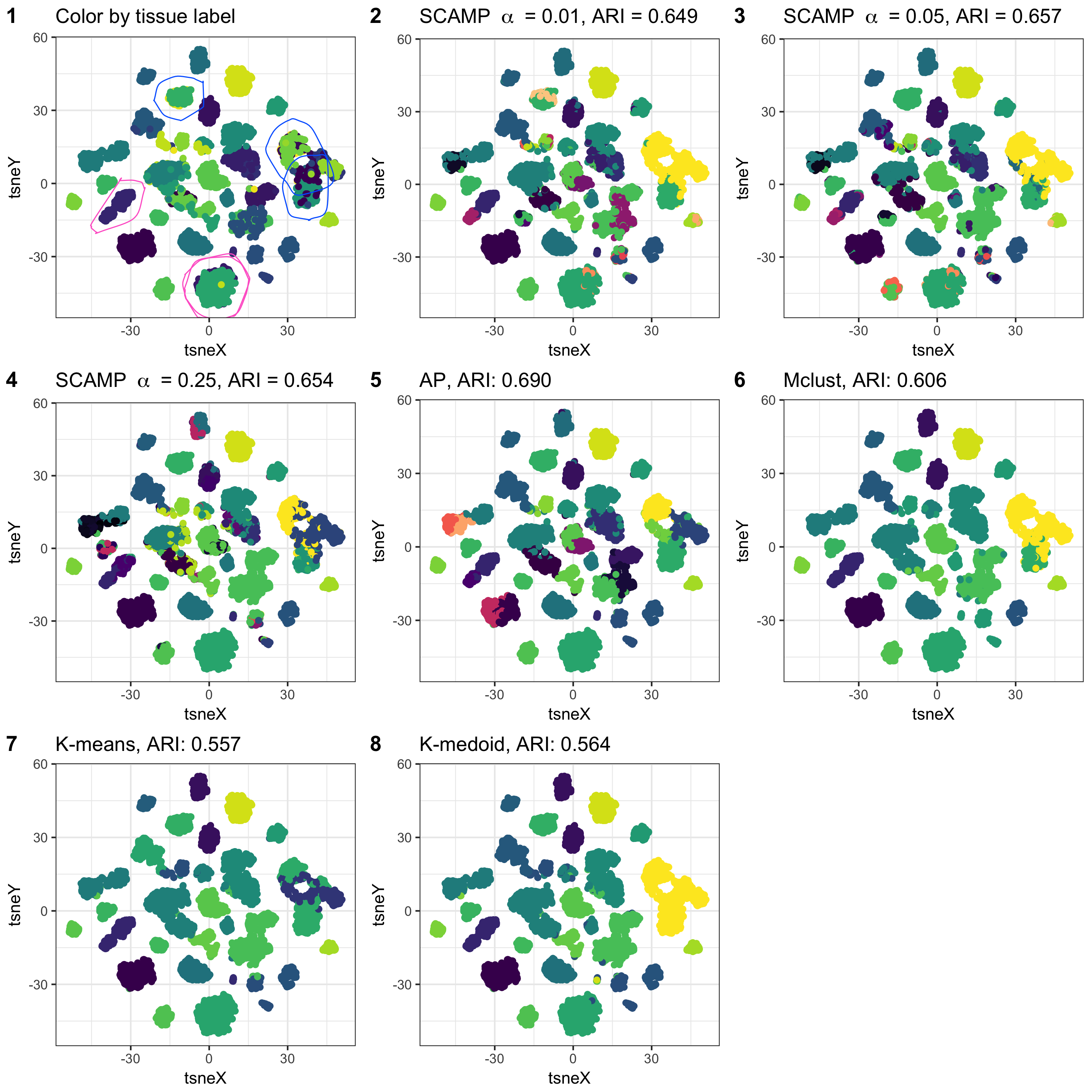}
  \caption[gtexTsne]{A t-SNE visualization of the GTEx data. 
  Panel 1 shows points colored by the $53$ GTEx tissue labels. 
    Generally, the ``islands'' in the t-SNE map correspond to a tissue type. 
    However, separation of ``islands'' occurs within tissues:
    the blue bounding boxes surround four ``islands'' with primary tissue label ``Brain'' 
    and thirteen tissue labels
    corresponding to specific regions of the brain. 
    The pink bounding boxes surround two ``islands'' with primary tissue label ``Skin''
    and three tissue labels "Cells - Transformed fibroblasts",
    "Skin - Not Sun Exposed (Suprapubic)",                    
    and "Skin - Sun Exposed (Lower leg)". 
    The remaining panels show the tissues colored according to SCAMP, affinity propagation,
    Mclust, k-means, and k-medoid clusterings of the data.
    Colors are manually selected to match a method's cluster with 
    the most similar ``true'' cluster (tissue label).
    When a method cluster has been matched to a ``true'' clusters, 
    subsequent method clusters that also match
    the same ``true'' cluster are colored by a separate palette.}\label{figure:gtexTsne}
\end{figure}

Numerical summaries of the clusterings are listed in the table in figure \ref{figure:gtexScores}.
The different methods perform comparably in terms of VI distance no matter
which set of labels are used as the truth.
k-medoid has the best performance in terms of ARI when the primary 
tissue labels are taken as the truth.  
Affinity propagation has the best performance in terms of ARI when the tissue labels are taken as
the truth.
Visualizing the data can help explain this difference: figure \ref{figure:gtexTsne} 
provides a visualization of these data through their t-SNE map.

In figure \ref{figure:gtexTsne}, we see both SCAMP 
and affinity propagation both have a tendency to partition primary tissue types into sub-clusters.
As the value of $\alpha$ increases from $0.01$ to $0.25$, the number of clusters found by SCAMP
increases. This can be seen in the t-SNE visualization, where homogeneous 
islands are split into overlapping clusters across the SCAMP panels.
It would require further analysis to determine if SCAMP and affinity propagation
are detecting groups of biological interest that are not detected by other methods (or each other).
However, since the goal of this section is to show how 
SCAMP can be productively applied to a modern dataset, we defer such analysis to subsequent work.

\section{Concluding Discussion}\label{section:conclusion} 

In this paper, we developed a new clustering algorithm called selective clustering annotated using 
modes of projections (SCAMP).
SCAMP relies heavily on the dip test of \cite{hartigan1985dip}; while developing SCAMP, we also 
developed an extension
of the dip test, described in section \ref{section:doubledip}, to test univariate distribution for 
multiple modes.
In section \ref{section:scamp}, we discussed the details of the SCAMP clustering algorithm.
In section \ref{section:casestudes}, we showed
that the SCAMP algorithm can used to produce interesting clusterings of many different types of data.

Over the course of the paper, we showed that SCAMP makes minimal distributional assumptions,
has tuning parameters that are relatively easy to set, can produce clusters with interpretable 
labels,
and its able to directly cluster datasets with a large number of observations.
We also discussed one of SCAMP's main limitations:
SCAMP can only detect clusters that are separable along the axes of measurement.
Additionally, we
proposed some work-arounds to this limitation: to apply SCAMP to the GTEx data set, we first
rotated the data through its principal components. 

As our analysis of the GTEx data shows, many clustering algorithms, including SCAMP,
can be used to generate useful clusterings of data matrices with a moderate number of observations.
However, many of these approaches also require the user to provide the number of clusters in the 
data as
a parameter. This requirement necessitates either a separate estimation step, a model comparison 
step, or choosing a large number
for this parameter and then merging clusters. By only requiring that the user set a level $\alpha$ 
for the dip test,
we think SCAMP has simplified the clustering task.

In terms of future research, we are interested in improving SCAMP's annotation step.
Currently SCAMP only annotates clusters relative to the columns in a data set with dip test 
p-values below level $\alpha$.
This is effectively a variable selection step. 
We think this step causes the SCAMP labels to omit some useful information.
For example, labels do not currently reflect a clusters position relative to a unimodal variables 
with wide shoulders;
it would be useful to improve the annotations to account for such a scenario.

We are also interested the consistency of phase 2 and phase 3 of the SCAMP algorithm.
While there are many hierarchical clustering algorithms, SCAMP's approach of recursively growing 
multiple partition trees and
then selecting leaves across different trees is, to our knowledge, new.
It is of interest to know if there are natural parametric conditions
(or, more generally, separation conditions) for the mixture components 
\eqref{scamp:cluster_distribution},
under which SCAMP can be shown to recover the number of components in the mixture and to
correctly assign observations to their generating component.

\section{Acknowledgements}

EG wishes to thank both K. K. Dey and V. Voillet for their separate discussions of the GTEx data. 
This work was funded by a grant from the Bill and Melinda Gates Foundation to RG [OPP1032317]
and a grant from the NIGMS to GF [R01 GM118417-01A1].


\newpage
\appendix

\section{Appendix} \label{section:appendix}

In these appendices, we provide proofs for section \ref{section:doubledip} 
and details for the separated mixture simulation in section \ref{section:casestudes}.

\subsection{Proofs} \label{section:appendixProof}

In this appendix, we provide proofs for lemmas and theorems found in section \ref{section:doubledip}.
\linebreak

\textbf{Proof of Lemma \ref{lem:dip_bound}}

\begin{proof}
   Let $q_\alpha$ denote the $\alpha$-quantile of the $F$-distribution, so that $F(q_{\alpha}-) \leq \alpha$ and
   $F(q_\alpha) \geq \alpha$. For a fixed value of $N$, select the $2N-1$ quantiles
  \begin{align}
    q_1\equiv q_{\frac{1}{2N}}, q_2 \equiv q_{\frac{2}{2N}},\ldots, q_{2N-1} \equiv q_{\frac{2N-1}{2N}}\ . \label{Fquantiles}
  \end{align}
  Suppose first that $F$ is continuous. This means the values \eqref{Fquantiles} are distinct.
  Begin by considering $H_1 \equiv F/F(q_1-)$. Per the discussion starting section 5 of \cite{dudleynotes2015}.,
  there exists a convex distribution function $C_1$ on $(-\infty,q_1)$ such that
  \[
    \sup_{x<q_1} |H_1-C_1| \leq 1/2 \ .
  \]
  Define $G_1 \equiv F(q_1-)\cdot C_1$. This is convex on $(-\infty,q1)$. Then
  \begin{align}
    \sup_{x < q_1} |F-G_1| = F(q_1-) \cdot \sup_{x < q_1} |\frac{F}{F(q_1-)}-C_1|\leq \frac{1}{4N}\ . \label{part1}
  \end{align}
  Next, define $H_N \equiv [F(x)-F(q_{2N-1})]/[1-F(q_{2N-1})]$. This gives a distribution function on $[q_{2N-1},\infty)$.
  There exists a concave distribution function $C_N$ on $[q_{2N-1},\infty)$ such that
  \[
    \sup_{x \geq q_{2N-1}} |H_N-C_N| \leq 1/2 \ .
  \]
  Define $G_N \equiv F(q_{2N-1})+(1-F(q_{2N-1}))C_N$, which is concave. Then
  \begin{align}
    \sup_{x \geq q_{2N-1}} |F-G_N| = (1-F(q_{2N-1})) \cdot \sup_{x > q_{2N-1}} |H_N-C_N|\leq \frac{1}{4N}\ . \label{part2}
  \end{align}
  Finally, for $j \in \left\{2,3,\ldots,2N-1\right\}$, define
  \[
    H_j \equiv \frac{[F(x)-F(q_{j-1})]}{F(q_{j}-)-F(q_{j-1})}
  \]
  on $[q_{j-1},q_j)$. The uniform d.f. on the interval $[q_{j-1},q_j]$, which we denote by $C_j$, has the trivial bound
  \[
    \sup_{q_{j-1} \leq x  \leq q_j} |H_j-C_j| \leq 1 \ .
  \]
  Define $G_j \equiv F(q_{j-1})+(F(q_j-)-F(q_{j-1}))C_j$. Then
  \begin{align}
    \sup_{q_{j-1} \leq x < q_{j}} |F-G_j| = (F(q_j-)-F(q_{j-1})) \sup_{q_{j-1} \leq x < q_{j}} |H_j-C_j|\leq F(q_j-)-F(q_{j-1}) \leq
    \frac{1}{2N}\ . \label{part3}
  \end{align}
  Finally define
  \begin{align}
    G(x) \equiv \sum_{j=1}^{2N} G_j 1_{[q_{j-1} \leq x < q_j)}\ , \label{mainapprox}
  \end{align}
  where we define $q_0 \equiv -\infty$ and $q_{2N} = \infty$. The linear components of $G(x)$ are both convex and concave,
  so by construction $G(x) \in \mathcal{U}_N$ and $\rho(F,G) \leq 1/2N$.

  If $F$ is not continuous, the quantiles \eqref{Fquantiles} may no longer be distinct.
  However, the construction used in the continuous case still essentially works. Define
  \[
    L \equiv \sup\left\{x | F(x) \leq  \frac{1}{2N}\right\}
  \]
  and
  \[
    U \equiv \inf\left\{x | F(x) \geq \frac{2N-1}{2N}\right\}\ .
  \]
  Then the constructions at \eqref{part1} and \eqref{part2} still produce the approximations on $(-\infty,L)$
  and $[U,\infty)$ that are smaller than $1/4N$.

  If $L=U$, then to conform with $\mathcal{U}_N$ we may proceed as follows: pick $\epsilon  > 0$ and
  apply the construction \eqref{part2} on the interval $[U+\epsilon,\infty)$. Then, divide the interval $[U,U+\epsilon)$
  equally according to the choice of $N$, and apply the linear approximation of \eqref{part3} to each component. Subsequently
  define $G(x)$ in \eqref{mainapprox} using these approximations.

  If $L < U$, then there are distinct quantile values \eqref{Fquantiles}. Repeated value in the list corespond to jump points,
  and we may repeat the method of the case $L=U$:
  insert linear approximations after jump points as needed to conform with $\mathcal{U}_N$.
\end{proof}

\textbf{Proof of Lemma \ref{lem:d_dd_equiv}}

\begin{proof}
  When $N = 1$, this is proved by Theorem 1 of \cite{hartigan1985dip}. For $N > 1$, the proof 
  of Hartigan and Hartigan extends naturally. Suppose $N > 1$.
  For $G \in \mathcal{U}_N$, define
  \[
    H(x) \equiv G(0)1_{[x < 0]}+G(x)1_{[0 \leq x \leq 1]} +G(1)1_{[x > 1]}\ .
  \]
  Recalling definition \ref{uclass}, suppose first $m_N < 0$. Then setting
  \[
    m_{H_1} =c_{H_1} = m_{H_2} = \ldots = m_{H_{N-1}} = c_{H_{N-1}} =  m_{H_N} \equiv 0
  \]
  places $H(x) \in \mathcal{V}_N$. If $m_1 > 1$, setting all modal and antimodal values to $1$ places $H(x) \in \mathcal{V}_N$
  Similarly, in the intermediate cases where $m_j < 0$, $c_j > 0$ , $c_k < 1$, and $m_{k+1} > 1$ with $j < k$, setting
  all modal and antimoal values below $0$ to $0$ and all values above $1$ to $1$ places $H(x) \in \mathcal{V}_N$. Thus,
  \[
    \rho(F,H) =  \sup_x |F(x)-H(x)| = \sup_{0 \leq x \leq 1} |F(x)-G(x)| \leq  \rho(F,G)\ .
  \]
  Hence $\rho(F,\mathcal{V}_N) \leq \rho(F,\mathcal{U}_N)$.

  Conversely, pick $G \in \mathcal{V}_N$. By Lemma \ref{lem:dip_bound}, it suffices to consider only $G$ such that
  $\rho(F,G) \leq 1/2N$. Such $G$ satify $0 \leq G(0) \leq 1/2N < 1-1/2N \leq G(1) \leq 1$.
  Define $\alpha \equiv 0 \vee G(0)$ , $\beta \equiv 1 \wedge G(1)$, and
  \[
    H(x) \equiv  \alpha 1_{[G < \alpha]} +G(x) 1_{[\alpha \leq G(x) \leq \beta]} + \beta 1_{[G > \beta]}\ .
  \]
  Since $G$ is non-decreasing on $[0,1]$ by definition, so too is $H$.
  Define $L \equiv \sup\left\{x\ |\ G(x) \leq \alpha\right\}$ and $U \equiv \inf\left\{x\ |\ G(x) \geq \beta\right\}$.
  For the values $m_1,c_1,\ldots,c_{N-1},m_N$ that fall below $\alpha$, set all of them to $G(L-)$ if the largest is a point of
  convexity. If the largest is a point of concavity, set that value to $G(L)$ (capturing the jump) and the remainder to $G(L-)$.
  Perform the analogous procedure for those points above $\beta$. Doing so, we see $H(x) \in \mathcal{V}_N$, and
  \[
    \rho(F,H) = \sup_{\alpha \leq G(x) \leq \beta}|F(x)-G(x)| \leq \rho(F,G)\ .
  \]
  Now for $a \geq 1$ define
  \[
    G_a(x) \equiv
    \begin{cases}
      GCM\left(H\cdot 1_{[x\ \geq -a]}, (-\infty,m_1)\right) \\ 
      H(x) \ \ \ \ \ \ \ \ \ \ \ \ \ \ \ \ \ \ \ \ \ \ \ \text{ for } x \in [m_1,m_N] \\
      LCM\left(H\cdot 1_{[x\ \leq a]}, (m_N,\infty)\right) \\
    \end{cases}\ .
  \]
  Here $G_a(x) \in \mathcal{U}_N$.  This extends the $G_a$ of Theorem 1 in \cite{hartigan1985dip}. The remainder
  of the proof follows the argument in Hartigan and Hartigan, with the expected notational changes. We include it for reference.

  The function $G_a$ is constant on $(-\infty,a]$, consists of a line segment on $[-a,x_1]$ for
  some $0 \leq x_1 \leq m_1$, and is equal to $H$ on $[x_1,m_1]$. Here we also have that as $a \nearrow \infty$,
  the slope of the linear segment tends to zero. Since $H$ is non-decreasing,
  \[
    \sup_{0 \leq x \leq m_1} |H(x)-G_a(x)| \rightarrow 0 \text{ as } a \nearrow \infty\ .
  \]
  By symmetry
  \[
    \sup_{m_N \leq x \leq 1} |H(x)-G_a(x)| \rightarrow 0 \text{ as } a \nearrow \infty\ .
  \]
  Since $G_a(x) = H(x)$ for $x \in [m_1,m_N]$, we conclude as $a \nearrow \infty$ that
  \[
    \rho(F,G_a) = \sup_{0 \leq x \leq 1 } |G_a(x) - F(x)| \rightarrow \sup_{0 \leq x \leq 1}|H(x) - F(x)| = \rho(F,H)\ .
  \]
  Thus for each $\epsilon > 0$ and $G\in\mathcal{V}_N$, there exists $G_a \in \mathcal{U}_N$ with
  \[
    \rho(F,G_a) \leq \rho(F,G) + \epsilon\ .
  \]
  Thus $\rho(F,\mathcal{U}_N) \leq \rho(F,\mathcal{V}_N)$.
\end{proof}

\textbf{Proof of Lemma \ref{lem:uniform_mix}}

\begin{proof}
  The argument of Theorem 2 in \cite{hartigan1985dip} applies here, with $D_N$ replacing $D$, and
  Lemma \ref{lem:d_dd_equiv} of this paper replacing Theorem 1 in \cite{hartigan1985dip}.
\end{proof}

\textbf{Proof of Theorem \ref{thm:conv_bb}}

\begin{proof}
  The argument of Theorem 3 in \cite{hartigan1985dip} applies here, with $D_N$ replacing $D$, and
  Lemma \ref{lem:uniform_mix} of this paper replacing Theorem 2 in \cite{hartigan1985dip}.
\end{proof}

\textbf{Proof of Theorem \ref{thm:dip_to_zero}}

\begin{proof}
  As noted in Section \ref{section:doubledip}, the claim essentially follows from Theorem 5 of \cite{hartigan1985dip}.
  We recapitulate the argument of Hartigan and Hartigan here, providing 
  details showing that the conditions \eqref{thm:dip_to_zero:c1} through
  \eqref{thm:dip_to_zero:c4} imply
  \begin{align}
    \sup_{-\infty \leq x_1 \leq x_2 \leq \infty} \frac{[F(x_2)-F(x_1)]^k}{|F(x_1)+F(x_2) - 2F(\overline{x})|} < \infty  \label{thm:dip_to_zero:main_bound}
  \end{align}
  so long as
  \[
    m_1,\ldots,m_N,c_1,\ldots,c_{N-1} \notin (x_1,x_2)\ . 
  \]
  Once demonstrated, the argument given in the second half of Theorem 5 in \cite{hartigan1985dip} proves the claim. We need to consider new details, since the additional peaks and valleys of multimodality mean we must consider
  some additional, but unsurprising, cases. 

  Without loss of generality, suppose $m_1 = 0$. To begin, if $x_1,x_2 < m_1=0$ or $x_1,x_2 > m_N$
  (and so \eqref{thm:dip_to_zero:c1} or \eqref{thm:dip_to_zero:c2} hold), \eqref{thm:dip_to_zero:main_bound} is proved by the argument given in the first half of Theorem 5 in \cite{hartigan1985dip}
  We now consider condition \eqref{thm:dip_to_zero:c3}. In the following analysis of cases, fix $j$ between $1 \leq j \leq N$.

  \textbf{Case 1}
  
  Pick $c_{j-1}+\epsilon_1 < x_1, x_2 < m_j -\epsilon_2$,  $F'(x) > 0$. Then,
  \[
    \frac{d}{dx} \log F'(x) > B \equiv B_{\epsilon_1} \wedge B_{\epsilon_2} > 0
  \]
  Thus
  \[
    \int_{x_1}^{x_2} \frac{d}{dx} \log F'(x) = \log\frac{F'(x_2)}{F'(x_1)} > B(x_2-x_1)\ ,
  \]
  and so
  \begin{align}
    \frac{F'(x_2)}{F'(x_1)} \geq \exp\left[B(x_2-x_1)\right] \ .  \label{thm:dip_to_zero:firstassumption}
  \end{align}
  This implies
  \begin{align}
  F(x_2)-F\left(\frac{x_1+x_2}{2}\right) \geq \left[F\left(\frac{x_1+x_2}{2}\right)-F(x_1)\right]\exp\left[B\left(\frac{x_2-x_1}{2}\right)\right]\ .
  \label{thm:dip_to_zero:immediateclaim}
  \end{align}
  To see that this is so, for $y \in [(x_1+x_2)/2,x_2] \equiv \mathcal{I}$, define the functions
  \begin{align}
    h(y) &\equiv F(y)\ , \nonumber \\
    g(y) &\equiv F\left(y-\frac{x_1+x_2}{2}+x_1\right) \ . \nonumber
  \end{align}
  Then by Cauchy's mean value theorem, there exists a $c \in ([x_1+x_2]/2,x_2)$ such that
  \[
    \frac{h(x_2) - h\left(\frac{x_1+x_2}{2}\right)}{g(x_2) - g\left(\frac{x_1+x_2}{2}\right)} = \frac{h'(c)}{g'(c)}\ .
  \]
  By their definition, this is equivalent to
  \[
    \frac{F(x_2) - F\left(\frac{x_1+x_2}{2}\right)}{F\left(\frac{x_1+x_2}{2}\right)-F(x_1)} = \frac{F'(c)}{F'\left(c-\frac{x_1+x_2}{2}+x_1\right)} \geq \exp\left(B\left[\frac{x_1+x_2}{2}-x_1\right]\right)
    =\exp\left(B\left[\frac{x_2-x_1}{2}\right]\right)\ ,
  \]
  with the inequality following by \eqref{thm:dip_to_zero:firstassumption}, justifying the claim. Let $\overline{x} \equiv (x_1+x_2)/2$ and $\delta \equiv (x_2-x_1)/2$. Observing $B\delta \geq 0$, we see \eqref{thm:dip_to_zero:immediateclaim} implies
  \[
    \frac{1}{1-[F(\overline{x})-F(x_1)] \bigg / [F(x_2)-F(\overline{x})]} \leq \frac {1}{1-\exp(-B \delta)} 
  \]
  and
  \[
    \left[1 + \frac{F(\overline{x})-F(x_1)}{F(x_2)-F(\overline{x})}\right]^k \leq \left[1+\exp\left(-B\delta\right)\right]^k \leq 2^k\ .
  \]
  Now if $B \delta \geq 1$,  we combine the preceding to find
  \[
    \frac{[F(x_2)-F(x_1)]^k}{|F(x_2)+F(x_1) - 2 F(\overline{x})|} \leq \frac{[F(x_2)-F(\overline{x})]^{k-1} 2^k}{1-\exp(-B \delta)} \leq \frac{2^k}{1-1/e} < \infty \ .
  \]
  Since $2/x > 1/[1-\exp(-x)]$ for $0 < x < 1$, if $0 < B\delta < 1$ we find
  \begin{align}
    \frac{[F(x_2)-F(x_1)]^k}{|F(x_2)+F(x_1) - 2 F(\overline{x})|} \leq \frac{[F(x_2)-F(\overline{x})]^{k-1} 2^k}{1-\exp(-B \delta)} \leq \frac{[F(x_2)-F(\overline{x})]^{k-1}2^{k+1}}{B\delta} \ . \label{thm:dip_to_zero:tmpbnd}
  \end{align}
  The mean value theorem gives for some $\eta \in ([x_1+x_2]/2,x_2)$ that
  \[
    F(x_2)-F(\overline{x})=F(x_2)-F\left(\frac{x_1+x_2}{2}\right) = \frac{x_2-x_1}{2}F'(\eta) = \delta F'(\eta)\ .
  \]
  At the mode $m_j$, we also have $F'(\eta) \leq F'(m_j)$. We thus continue from \eqref{thm:dip_to_zero:tmpbnd} to see
  \[
    \frac{[F(x_2)-F(x_1)]^k}{|F(x_2)+F(x_1) - 2 F(\overline{x})|}  \leq \frac{\delta^{k-2}[F'(m_j)]^{k-1}2^{k+1}}{B}  < \frac{[F'(m_j)]^{k-1}2^{k+1}}{B^{k-1}}  < \infty \ .
  \]

  \textbf{Case 2}
  
  Since we assume there is a $k \geq 2$ for which $F^{(k)}(m_j) \neq 0$, pick $\epsilon$ small
  enough that the sign of $F^{(k)}(x)$ does not change for $x \in [m_j-\epsilon,m_j]$
  and so that $F^{(x)}(x) \neq 0$. 
  Now suppose $m_j-\epsilon < x_1 < x_2 < m_j$.   
  Two applications of Taylor's theorem give
  \[
    F(x_2)+F(x_1)-2F(\overline{x}) = \frac{(x_2-x_1)^2}{8}\left[F''(y_1)+F''(y_2)\right]\ .
  \]
  with  $y_1 \in (x_1,\overline{x})$ and $y_2 \in (\overline{x},x_2)$. 
  If, excepting the first derivative, $F^{(k)}(m_j)$ is the first non-zero derivative at $m_j$, the Taylor polynomial of $F''(y_1)$ is
  \[
    P_{k-3}(y_1-m_j) = F''(m_j)+F'''(m_j)(y_1-m_j)+\dots+ \frac{F^{(k-1)}(m_j)}{(k-3)!} (y_1-m_j)^{k-3} = 0\ ,
  \]
  with remainder 
  \[
    R_{k-3}(y_1-m_j) = \frac{F^{(k)}(\eta_{y_1})}{(k-2)!} (y_1-m_j)^{k-2} \ .
  \]      
  The analogous expansion of $F''(y_2)$ at $m_j$ gives
  \begin{align}
  |F(x_2)+F(x_1)-2F(\overline{x})| &=
  \frac{(x_2-x_1)^2}{8(k-2)!}
  \cdot |F^{(k)}(\eta_{y_1})(y_1-m_j)^{k-2} + F^{(k)}(\eta_{y_2})(y_2-m_j)^{k-2}|
  \nonumber  \\
  &\geq
  \frac{(x_2-x_1)^2}{8(k-2)!}
  \cdot |F^{(k)}(\eta_{k})(y_1-m_j)^{k-2} + F^{(k)}(\eta_{k})(y_2-m_j)^{k-2}|
  \label{appendix:etaChoice} \\
  &\geq
   \frac{(x_2-x_1)^2 \cdot |F^{(k)}(\eta_{k})|\cdot |(m_j-\overline{x})^{k-2}|}{8(k-2)!}
    \nonumber
  \end{align}
  with $\eta_k \in \left\{\eta_{y_1},\eta_{y_2}\right\}$ chosen as a function of $k$ to minimize
  the sum \eqref{appendix:etaChoice}.
  Again using the mean value theorem, we have
  \[
    F(x_2)-F(x_1) = (x_2-x_1)F'(\zeta) \leq (x_2-x_1)F'(m_j)\ .
  \]
  Since $(x_2-x_1)/(m_j-\overline{x}) \leq 2$, we find
  \[
    \frac{[F(x_2)-F(x_1)]^k}{|F(x_2)+F(x_1)-2F(\overline{x})|} \leq \frac{8(x_2-x_1)^{k-2}[F'(m_j)]^k \cdot (k-2)!}{|F^{(k)}(\eta_k)|\cdot |m_j-\overline{x}|^{k-2}} \leq \frac{ 2^{k+1}[F'(m_j)]^k\cdot (k-2)!}{|F^{(k)}(\eta_k)|} < \infty \ .
  \]

  \textbf{Case 3}

  Next suppose $c_{j-1} < x_1 < x_2 < c_{j-1}+\epsilon$. A similar analysis to that given in case 2 shows
  \begin{align}
    |F(x_2)+F(x_1)-2F(\overline{x})|
    \geq \frac{(x_2-x_1)^2|F^{(k)}(\eta_k)|\cdot |(\overline{x}-c_{j-1})^{k-2}|}{8(k-2)!}\ . \nonumber 
  \end{align}
  However, the mean value theorem and increase in $x$ still gives
  \[
    F(x_2)-F(x_1) = (x_2-x_1)F'(\zeta) \leq (x_2-x_1)F'(m_j)\ ,
  \]
  and so the analogous bound holds.

  \textbf{Case 4}

  Now suppose $c_{j-1}+\epsilon_1 < x_1 \leq m_j-\epsilon_2 < m_j-\eta \leq x_2 < m_j$, and that $F(m_j)-F(m_j-\eta) \leq F(m_j-\eta) - F(m_j-\epsilon_2)$. The assumptions imply:
  \[
    F(x_2)-F(m_j-\eta) \leq F(m_j)-F(m_j-\eta) \leq F(m_j-\eta)-F(m_j-\epsilon_2) \leq F(m_j-\eta)-F(x_1)\ .
  \]
  From this it follows
  \[
    F(m_j-\eta) + [F(m_j-\eta) - F(x_1) ] - F(x_2) \geq F(m_j-\eta) + [F(m_j) - F(m_j-\eta)] - F(x_2) = F(m_j)-F(x_2) \geq 0\ .
  \]
  Re-arranging, this implies
  \begin{align}
    F(x_2)-F(x_1) \leq 2[F(m_j-\eta)-F(x_1)]\ . \label{thm:dip_to_zero:doubling}
  \end{align}
  Re-arranging and using convexity, it also implies
  \[
    F(m_j-\eta) \geq \frac{F(x_1)+F(x_2)}{2} \geq F(\overline{x})\ . 
  \]
  Now define
  \[
    h_{x_1}(t) \equiv F(t) + F(x_1) - 2F\left(\frac{t+x_1}{2}\right)\ ,
  \]
  with $t \in (x_1, m_j)$. Recall $F$ is convex and non-decreasing on $[c_{j-1}+\epsilon_1,m_j]$. Differentiating, we thus see
  \[
    h_{x_1}'(t) = F'(t)- F'\left(\frac{t+x_1}{2}\right)  > 0\ .
  \]
  Since $m_j-\eta < x_2$, we thus have
  \[
    h_{x_1}(m_j-\eta) \leq h_{x_1}(x_2) \Longleftrightarrow F(m_j-\eta) + F(x_1) - 2F\left(\frac{m_j-\eta+x_1}{2}\right) \leq F(x_2) + F(x_1) - 2F\left(\frac{x_2+x_1}{2}\right)\ .
  \]
  This is equivalent to
  \[
    \frac{1}{F(x_2) + F(x_1) - 2F\left(\frac{x_2+x_1}{2}\right)} \leq \frac{1}{F(\eta) + F(x_1) - 2F\left(\frac{m_j-\eta+x_1}{2}\right)}\ .
  \]
  Combining this with \eqref{thm:dip_to_zero:doubling}, we infer
  \[
    \frac{[F(x_2)-F(x_1)]^k}{F(x_2) + F(x_1) - 2F\left(\frac{x_2+x_1}{2}\right)} \leq \frac{2^k[F(m_j-\eta)-F(x_1)]^k}{F(m_j-\eta) + F(x_1) - 2F\left(\frac{m_j-\eta+x_1}{2}\right)} \leq 2^kC_\eta < \infty\ ,
  \]
  with the penultimate inequality following by Case 1.

  \textbf{Case 5}
  
   Suppose now that $c_{j-1} < x_1 \leq c_{j-1}+\eta < c_{j-1}+\epsilon_1 \leq x_2 \leq m_j-\epsilon_2$, and that $F(c_{j-1}+\eta)-F(c_{j-1}) \leq F(c_{j+1}+\epsilon_1) - F(c_{j-1}+\eta)$. Similar to Case 4,
  \[
    F(c_{j-1}+\eta)-F(x_1) \leq F(c_{j-1}+\eta)-F(c_{j-1}) \leq F(c_{j-1}+\epsilon_1)-F(c_{j-1}+\eta) \leq F(x_2)-F(c_{j-1}+\eta)\ .
  \]
  From this follows
  \begin{align}
    F(x_2)-F(x_1) \leq 2[F(x_2)-F(c_{j+1}+\eta)]\ . \label{thm:dip_to_zero:doubling_again}
  \end{align}
  A symmetric argument with Case 4 gives
  \[
    \frac{[F(x_2)-F(x_1)]^k}{F(x_2) + F(x_1) - 2F\left(\frac{x_2+x_1}{2}\right)} \leq \frac{2^k[F(x_2)-F(c_{j-1}+\eta)]^k}{F(x_2) + F(c_{j-1}+\eta) - 2F\left(\frac{c_{j-1}+\eta+x_2}{2}\right)} \leq 2^k\ C_\eta < \infty\ ,
  \]

  \textbf{Case 6}
  
  Finally, suppose
  \[
    c_{j-1} < x_1 \leq c_{j-1}+\eta_1 \leq  c_{j-1}+\epsilon_1  < m_j-\epsilon_2 \leq m_j - \eta_2 \leq x_2 < m_j \ ,
  \]
  with both
  \[
    F(c_{j-1}+\eta_1)-F(c_{j-1}) \leq F(c_{j+1}+\epsilon_1) - F(c_{j-1}+\eta_1)
  \]
  and
  \[
    F(m_j)-F(m_j-\eta_2) \leq F(m_j-\eta_2) - F(m_j-\epsilon_2)\ .
  \]
  Then
  \[
    F(c_{j-1}+\eta_1)-F(x_1) \leq F(c_{j-1}+\eta_1)-F(c_{j-1}) \leq F(c_{j+1}+\epsilon_1) - F(c_{j-1}+\eta_1) \leq F(x_2) - F(c_{j-1}+\eta_1)
  \]
  and
  \[
    F(x_2)-F(m_j-\eta_2) \leq F(m_j)-F(m_j-\eta_2) \leq F(m_j-\eta_2)-F(m_j-\epsilon_2) \leq F(m_j-\eta_2)-F(x_1)\ .
  \]
  Jointly these imply
  \[
    F(x_2)-F(x_1) \leq 2[F(m_j-\eta_2)-F(c_{j-1}+\eta_1)]\ .
  \]
  Combining the argument of cases 4 and 5 gives the bound
  \[
    \frac{[F(x_2)-F(x_1)]^k}{F(x_2) + F(x_1) - 2F\left(\frac{x_2+x_1}{2}\right)} \leq \frac{2^k[F(m_j-\eta_2)-F(c_{j-1}+\eta_1)]^k}{F(m_j-\eta_2) + F(c_{j-1}+\eta_1) - 2F\left(\frac{c_{j-1}+m_j+\eta_1-\eta_2}{2}\right)} \leq 2^k\ C_{\eta_1,\eta_2} < \infty\ .
  \]

  \textbf{Conclusion}

  Taken together, cases $1-6$ show condition \eqref{thm:dip_to_zero:c3} implies \eqref{thm:dip_to_zero:main_bound}. Symmetric arguments imply when \eqref{thm:dip_to_zero:c4} holds,
  \eqref{thm:dip_to_zero:main_bound} holds. We thus see \eqref{thm:dip_to_zero:main_bound} follows from the given assumptions.
  
  To show $\sqrt{n}D_N(F_n) \rightarrow_p 0$, construct $G_n$ as follows: let $G_n$ be the LCM of $F_n$ in $[-\infty,m_1]$, the GCM of $F_n$ on $[m_N,\infty]$, the LCM of
  $F_n$ on intervals $[c_{j-1},m_j]$, and the GCM of $F_n$ on intervals $[m_j,c_j]$. Then $G_n \in \mathcal{U}_N$. 
  Considering maximal linear components of $G_n$ that occurs between the atoms $m_1,\ldots,m_N,c_1,\ldots,c_{N-1}$, such that $G_n=F_n$ at the end-points. The argument of Hartigan and Hartigan in Theorem 5 of \cite{hartigan1985dip}
  show these intervals are of length $o_P(1)$. Convergence in probability follows.
\end{proof}

\subsection{Separated Mixture Settings} \label{section:appendixSettings}


This section of the appendix describes our settings for the separated mixtures simulation.
\begin{table}[ht] 
\centering
\begin{tabular}{lrr}
  \hline
  & Default & Elaboration \\
  \hline
  Second Mixture Dimension: &   0 &  20  \\
  Second Mixture Mean Different: &   False &  True  \\
  Noise Component Dimension: &   0 &  20 \\
  Transform Coordinates: & False & True \\
  Sampling Regime: & Gaussian & T plus Gaussian \\
  \hline
\end{tabular}
\caption{Simulation Parameter}\label{simulation:mixtureSettings}
\end{table}
When all parameters are set to their default, the mixture components of \eqref{section4:mixtureDesc} are all multivariate Gaussian.
The mean vector of each component is randomly selected so its entries lie in $\left\{0,6\right\}$.
The variance-covariance matrix is randomly selected from the space of positive definite matrices with variance bounded above by 3.

We now describe how parameter settings affect the simulation.
When the ``Second Mixture Dimension'' parameter is set to $20$, a random sample from a 
\textit{second mixture} of either size $3000$ or $30000$ is taken. The number of rows 
depends on the simulation scenario.
This leads to a data matrix of size $3000 \times 40$ or $30000 \times 40$, depending on the 
simulation scenario.
The clustering ground truth is taken to be the cross-product of the labels of the mixture
components which produces a row of the data matrix.
This creates $17$ distinct clusters ranging in size between $125$ and $375$ observations.

The ``Second Mixture Mean Different'' parameter only affects the simulation if the ``Second Mixture
Dimension'' parameter is set to $20$.
If the ``Second Mixture Mean Different'' parameter is set to ``False'', the mean 
vectors of components in the simulated mixture have entries randomly chosen from
$\left\{0,6\right\}$.
If the ``Second Mixture Mean Different'' parameter is set to ``True'', the
mean vectors of components in the simulated mixture have entries randomly chosen from
$\left\{0,3,6\right\}$.

If the ``Noise Component Dimension'' parameter is set to $20$, a sample of $3000$ 
or $30000$ observations is taken from a multivariate t distribution with $5$ degrees of freedom.
The non-centrality parameter is set to $2.5$ for each component.
The scale matrix sampled from the space of positive definite matrices with diagonal elements 
bounded above by 3.
This sample is adjoined to the samples from the mixture distribution(s). It is mean to represent 
the case when measurements are taken on variables unrelated to the signal of interest.
Since the sample is from a single distribution, it does not affect the clustering ground truth.

If the ``Transform Coordinates'' parameter is set to ``True'', rows sampled from the mixtures are taken through the following maps:
\begin{align}
  f_1(x) &= \left(\sqrt{|x_1|},\ldots,\sqrt{|x_p|}\right)\ . \nonumber \\
  f_2(x) &= \left(\exp(x_1),\ldots,\exp(x_p)\right)\ . \nonumber \\
  f_3(x) &= \left(x_1^2,\ldots,x_p^2\right)\ . \nonumber \\
  f_4(x) &= \left(x_1,\ldots,x_p\right)\ . \label{scamp:sim_settings} 
\end{align}
The maps are sequentially applied to the components of the mixture, starting with $f_1$.
Once all four maps have been applied to different mixture components, the process repeats
from $f_1$, until all clusters have been taken through a map.

Finally, if the ``Sampling Regime'' parameter is set to ``T plus Gaussian'', components of the mixture alternate between multivariate normal and multivariate t with $5$ degrees of freedom.
For the multivariate t, the non-centrality parameter and scale matrix are sampled from the same space as its Gaussian counterpart.
If the ``Sampling Regime'' parameter is set to ``Gaussian'', components of the mixture are all initially samples from multivariate normal.

\subsubsection{Scenario 1: Moderate sample size, moderate number of mixture components}

To simulate this scenario, we begin by taking a sample of size $3000$ from a mixture in $20$ dimensions.
The mixture components have the following sizes, and are sampled in the stated order:
\begin{align}
  \text{Mixture Component Sizes } \equiv\ \left\{1000,500,250,250,250,250,125,125,125,125\right\}\ .\label{section4:mixtureDesc}
\end{align}
In this setting when the ``Second Mixture Dimension'' parameter is set to $20$, a random sample from a \textit{second mixture} of size $3000$ is taken.
The mixture components have the following sizes; samples are taken the stated order:
\begin{align}
  \text{Second Mixture Component Sizes } \equiv\ \left\{250,125,250,125,125,250,500,125,250,1000\right\}\ .\label{section4:mixtureDesc2BigSimple}
\end{align}
As before, this creates $17$ distinct clusters.

\subsubsection{Scenario 2: Large sample size, moderate number of mixture components}

The mixture components have the following sizes, and are sampled in the stated order:
\begin{align}
  \text{Mixture Component Sizes } \equiv\ \left\{10000,5000,2500,2500,2500,2500,1250,1250,1250,1250\right\}\ .\label{section4:mixtureDescBigSimple}
\end{align}
In this setting when the ``Second Mixture Dimension'' parameter is set to $20$, a random sample from a \textit{second mixture} of size $30000$ is taken.
The mixture components have the following sizes; samples are taken the stated order:
\begin{align}
  \text{Second Mixture Component Sizes } \equiv\ \left\{2500,1250,2500,1250,1250,2500,5000,1250,2500,10000\right\}\ .\label{section4:mixtureDesc2BigSimple}
\end{align}
As before, this creates $17$ distinct clusters.

\subsubsection{Scenario 3: Large sample size, moderate number of mixture components}

The mixture components have the following sizes, and are sampled in the stated order:
\begin{align}
  \text{Large Sample Mixture Component Sizes } \equiv\
  \left\{
  \begin{matrix}
    &10000,5000,2500,2500,1250,1250\\
    &1250,1250,500,500,500,500\\
    &250,250,250,250,250,250\\
    &250,250,100,100,100,100\\
    &100,100,100,100,100,100\\
  \end{matrix}
  \right\}\ .\label{section4:mixtureDescBig}
\end{align}
In this setting when the ``Second Mixture Dimension'' parameter is set to $20$, a random sample from a \textit{second mixture} of size $30000$ is taken.
The mixture components have the following sizes; samples are taken the stated order:
\begin{align}
  \text{Large Sample Second Mixture Component Sizes } \equiv\
  \left\{
  \begin{matrix}
    &250,100,250,100,100,250\\
    &500,100,100,100,1250,250\\
    &500,100,250,1250,100,500\\  
    &5000,250,10000,2500,250,1250\\
    &100,100,500,2500,1250,250\\
  \end{matrix}
  \right\}\ .\label{section4:mixtureDescBig2}
\end{align}
When the mixtures \eqref{section4:mixtureDescBig2} and \eqref{section4:mixtureDescBig} are combined, they 
produce a mixture with 56 components, with sample sizes
oranging between $50$ and $3950$ observations.

\bibliographystyle{authordate1}
\bibliography{finalSources}

\end{document}